\documentclass{article}

\usepackage[final]{neurips_2025}

\usepackage[utf8]{inputenc} % allow utf-8 input
\usepackage[T1]{fontenc}    % use 8-bit T1 fonts
\usepackage{hyperref}       % hyperlinks
\usepackage{url}            % simple URL typesetting
\usepackage{booktabs}       % professional-quality tables
\usepackage{amsfonts}       % blackboard math symbols
\usepackage{nicefrac}       % compact symbols for 1/2, etc.
\usepackage{microtype}      % microtypography
\usepackage{xcolor}         % colors
\usepackage{makecell}

\usepackage{amsmath}
\usepackage{amssymb}
\usepackage{graphicx}
\usepackage{subcaption}
\usepackage{wrapfig}  % Add this line for fig right part

\usepackage[utf8]{inputenc}
\usepackage[T1]{fontenc}
\usepackage{lipsum} % 只是为了生成一些示例文字
\usepackage{xcolor} % 用于颜色定义
\usepackage{tcolorbox}
\usepackage{svg}

\usepackage{multirow}
\definecolor{rosepink}{RGB}{255, 102, 204}

\title{%
  \raisebox{-0.2\height}{\includegraphics[width=20pt]{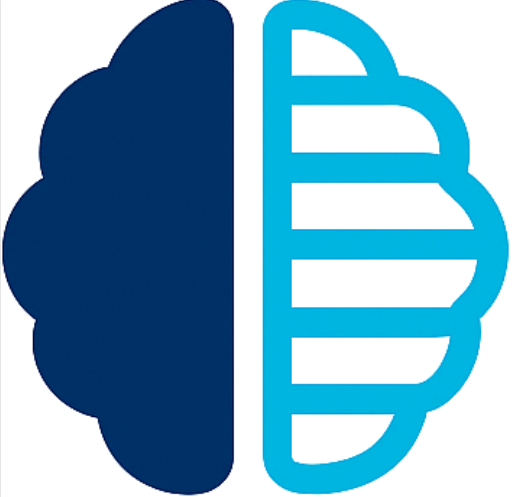}}~
  Aux-Think: Exploring Reasoning Strategies for Data-Efficient Vision-Language Navigation
}

\author{
    Shuo Wang$^{1,3}$\thanks{This work was done while Shuo Wang was a Research Intern with Horizon Robotics.},\; 
    Yongcai Wang$^{1}$\thanks{Corresponding authors.},\;
    Wanting Li$^1$\;,\;
    Xudong Cai$^1$,\; \\
    \textbf{Yucheng Wang}$^3$\thanks{Project leader.},\;
    \textbf{Maiyue Chen}$^3$,\;
    \textbf{Kaihui Wang}$^3$,\;
    \textbf{Zhizhong Su}$^3$,\;
    \textbf{Deying Li}$^1$,\;
    \textbf{Zhaoxin Fan}$^{2}$\footnotemark[2]\\[2mm]
    $^1$Renmin University of China,
    $^2$Beijing Advanced Innovation Center for Future Blockchain\\ and Privacy Computing,
    $^3$Horizon Robotics \\[2mm]
    \textcolor{rosepink}{\texttt{\url{https://horizonrobotics.github.io/robot_lab/aux-think}}}
}
\begin{document}

\maketitle

\vspace{-0.5cm}
\begin{figure}[htbp]
    \centering 
    \begin{subfigure}[t]{0.4\textwidth} 
        \centering % Center content within the subfigure
        \includegraphics[width=0.99\linewidth]{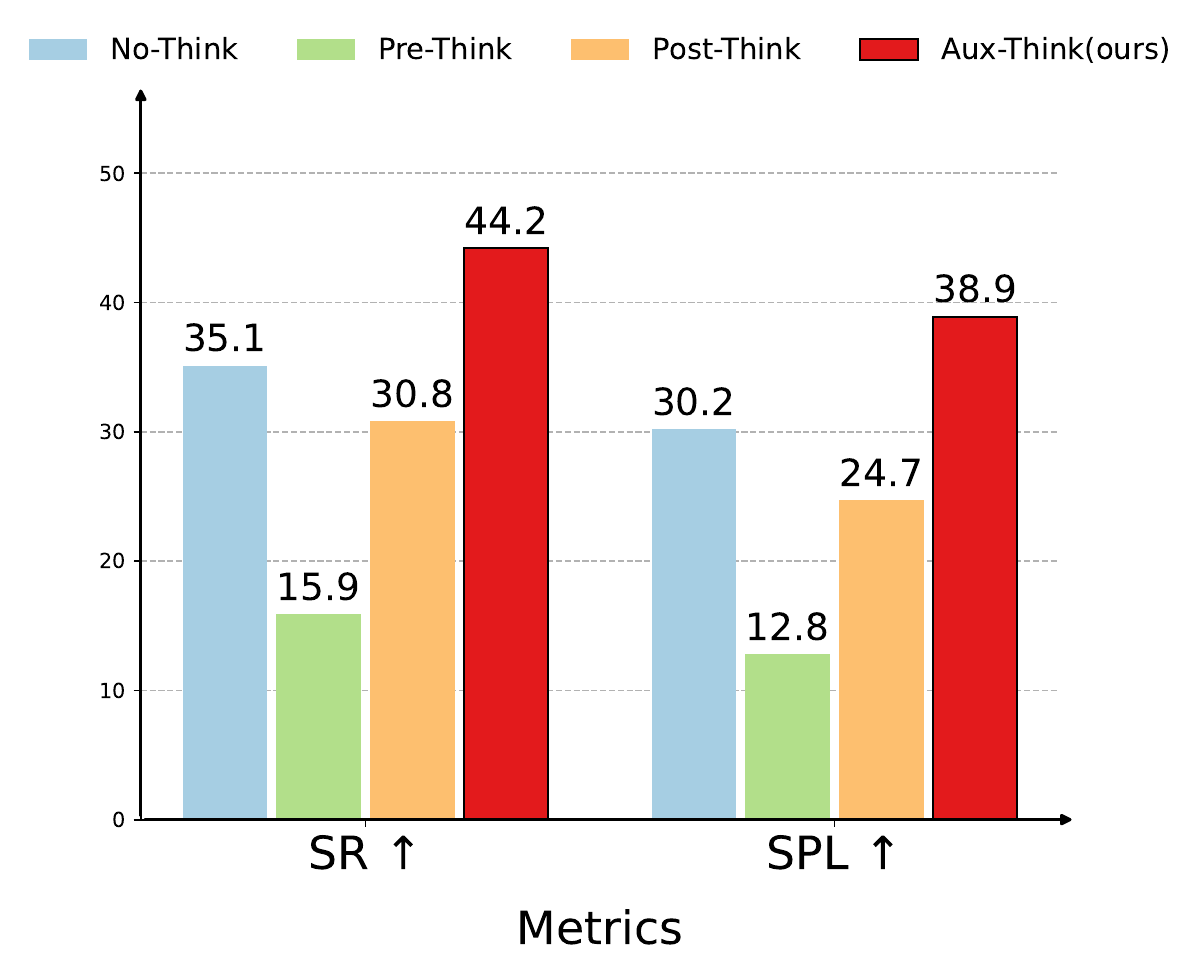}
        \caption{Comparison of navigation performance of different reasoning strategies, which are only trained on R2R-CoT-320k without receding-horizon action planning. The proposed Aux-Think (Ours) method consistently outperforms other reasoning strategies.}
        \label{fig:1a} % Changed label to reflect it's part (a)
    \end{subfigure}
    \hspace{0.05\textwidth} % Adjust spacing between subfigures if needed
    % Subfigure (b)
    \begin{subfigure}[t]{0.4\textwidth} % Or adjust width as needed, [t] aligns tops
        \centering % Center content within the subfigure
        \includegraphics[width=1\linewidth]{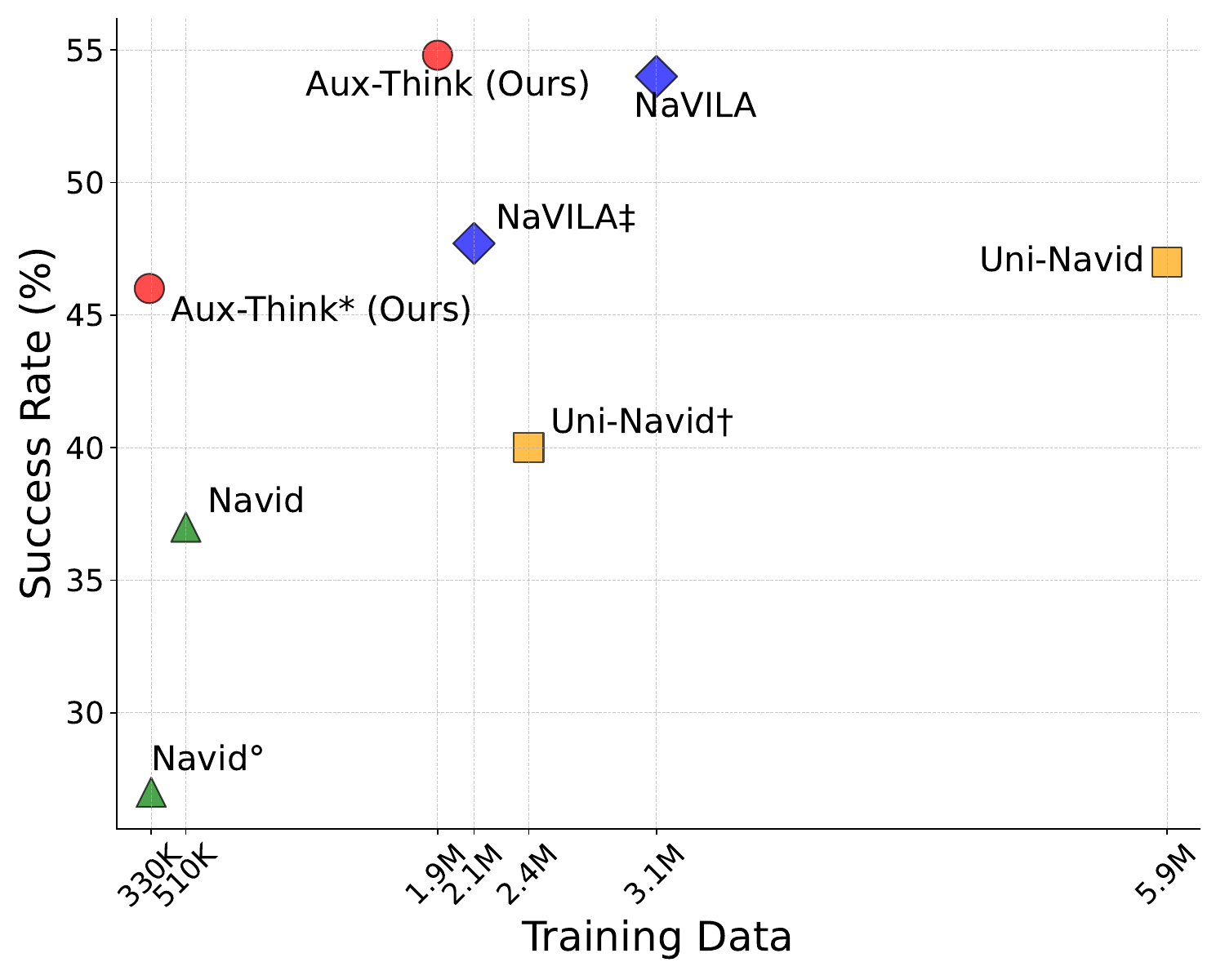}
        \caption{Aux-Think (Ours) is Pareto-optimal in data efficiency and success rate.
        Variants: Aux-Think*: only trained on R2R-CoT-320k. Navid$^\circ$ : no DAgger~\cite{ross2011reduction} and instruction data. Uni-Navid$\dagger$: no VQA data. NaVILA‡: no Internet data.} % Note: "method name" seems like a placeholder here
        \label{fig:1b} % Added a label for part (b)
    \end{subfigure}
    \caption{Aux-Think outperforms alternative reasoning approaches in navigation tasks (a) and achieves a favorable trade-off between data usage and success rate (b).} % Add an overall caption for the figure
    \label{fig:overall_comparison}
\end{figure}

\begin{abstract}

Vision-Language Navigation (VLN) is a critical task for developing embodied agents that can follow natural language instructions to navigate in complex real-world environments. Recent advances driven by large pretrained models have significantly improved generalization and instruction grounding compared to traditional approaches. However, reasoning strategies in this task remain underexplored. Navigation is action-centric and long-horizon, while Chain-of-Thought (CoT) reasoning has mainly shown success in static tasks such as visual question answering. To address this gap, we conduct the first systematic evaluation of reasoning strategies, including No-Think (direct action prediction), Pre-Think (reasoning before action), and Post-Think (reasoning after action). Surprisingly, our findings reveal a Test-time Reasoning Collapse issue, where reasoning during testing degrades navigation accuracy, highlighting the challenges of integrating reasoning into embodied navigation. Based on this insight, we propose Aux-Think, a framework that trains models to internalize structured reasoning patterns via CoT supervision, while predicting actions directly without explicit reasoning at test time. To support this framework, we release R2R-CoT-320k, the first Chain-of-Thought annotated dataset for VLN. Extensive experiments show that Aux-Think substantially reduces training effort and achieves state-of-the-art performance on success rate.
% Project page: \url{https://horizonrobotics.github.io/robot_lab/aux-think}.

\end{abstract}

\section{Introduction}
\label{sec:intro}

Vision-and-Language Navigation (VLN) \cite{wu2024vision,anderson2018vision,gu2022vision} represents a groundbreaking step towards enabling robots to understand natural language instructions and navigate complex, unfamiliar environments. As a foundational capability for embodied AI systems, VLN bridges the gap between perception and action, empowering robots to seamlessly interact with the real world. In particular, Vision-Language Navigation in continuous environments (VLN-CE) \cite{hong2022bridging,wang2023gridmm,an2022bevbert} has emerged as a critical research focus, pushing the boundaries of autonomy and adaptability in dynamic, real-world scenarios.

Traditional Vision-and-Language Navigation methods often rely on waypoint predictors \cite{krantz2021waypoint,hong2022bridging,krantz2022sim} or topology maps \cite{zubair2021sasra,chen2021topological,chen2022weakly,wang2025mambavo} but struggle with generalization and the sim-to-real gap. With the advancements in Large Language Models (LLMs) \cite{yang2024qwen2, touvron2023llama} and Vision-Language Models (VLMs) \cite{bai2025qwen2,liu2024nvila,lin2023video}, recent studies \cite{zhang2024navid,zhang2024uni,cheng2024navila} shift toward action prediction via supervised fine-tuning on paired videos and instructions. %, achieving strong performance by scaling data. 
Despite these advancements, most efforts emphasize training strategies \cite{zhu2020vision,zeng2024poliformer}, data organization \cite{han2024roomtour3d}, or model architecture \cite{zheng2024towards} for VLN.
Chain-of-Thought (CoT), which explicitly generates intermediate reasoning before producing final answers \cite{wei2022chain}, has shown success in enhancing reasoning capabilities across various LLM- and VLM-driven tasks, like video understanding \cite{feng2025video} and tool usage \cite{lu2025ui}. However, its application to VLN remains unexplored.

\begin{figure}[htbp]
    \centering
    \includegraphics[width=1\linewidth]{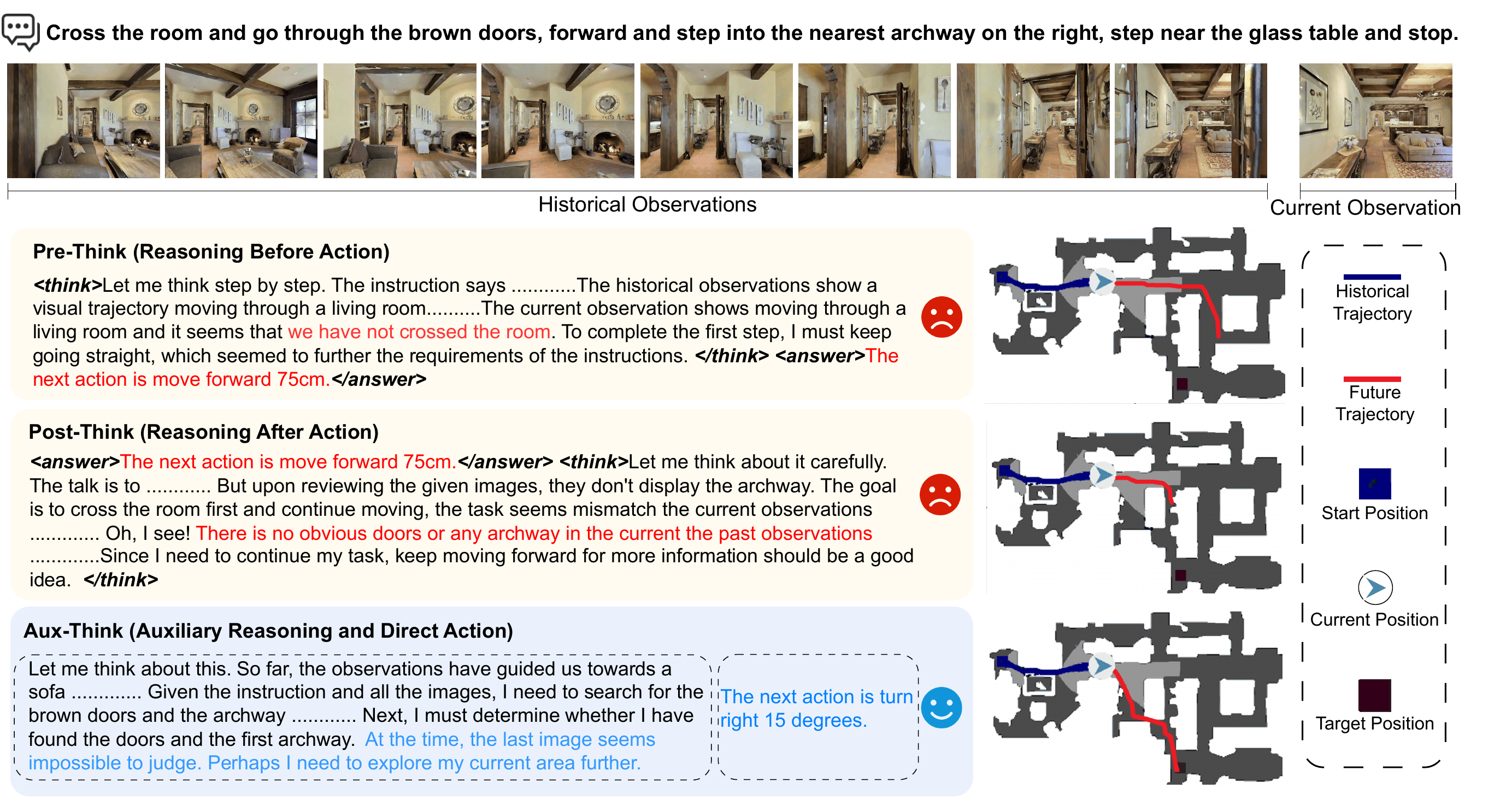}
    \caption{Illustration of CoT and Action Prediction Results Using Different Reasoning Strategies. Pre-Think generates incorrect actions (e.g., “move forward 75cm”) due to flawed CoT reasoning, such as “we have not crossed the room,” leading to significant trajectory deviation. Post-Think, which builds on Pre-Think’s output, inherits similar reasoning errors (e.g., “no obvious door or archway”) and makes the same wrong prediction. In contrast, Aux-Think correctly predicts “turn right 15 degrees” and follows a trajectory aligned with the ground truth. While Aux-Think does not rely on CoT during testing, it can optionally produce CoT via prompt switching—yet its action prediction remains accurate even when the generated CoT is of moderate quality. This highlights Aux-Think’s robustness to imperfect reasoning and its superior reliability in action prediction.}
    \label{fig:visulization}
\end{figure}

Motivated by this gap, we present the first systematic study of reasoning strategies in VLN, comparing three paradigms:
(1) No-Think, which predicts actions without explicit reasoning; (2) Pre-Think \cite{zhang2024chain}, which performs CoT before action selection; and (3) Post-Think \cite{li2025think}, which reasons after action prediction. 
\textbf{Our key findings reveal a phenomenon we term ``Test-time Reasoning Collapse'' (TRC)}: 
Introducing CoT via Pre-Think or Post-Think consistently harms navigation performance (Fig. \ref{fig:1a}).
CoT errors or hallucinations during testing result in incorrect actions, as shown in Fig. \ref{fig:visulization}.
We attribute this to a training–testing mismatch: CoT is only trained on optimal (oracle) trajectories, but for testing, agents often enter unfamiliar, off-distribution states where reasoning fails, leading to error accumulation along the trajectory and cascading navigation failures.
Even with DAgger, coverage of off-distribution states is limited and CoT supervision remains biased toward the optimal region.
This phenomenon highlights a fundamental limitation of multi-turn explicit reasoning in dynamic, partially observable environments, unlike the single-turn static tasks such as VQA or image captioning \cite{antol2015vqa,wang2024videocot}.

To address the TRC issue, we propose \textbf{Aux-Think}, inspired by the dual-process theory of human learning \cite{evans2003two}: during training, humans often rely on explicit reasoning to understand principles, but during execution, they focus on actions without consciously recalling those principles, much like a driver who no longer recites traffic rules while driving. Similarly, Aux-Think uses CoT as an auxiliary signal during training to guide the model in internalizing reasoning patterns. At testing time, the model no longer generates explicit reasoning, but instead directly predicts actions based on the internalized reasoning learned during training. This separation between learning and execution improves decision focus, reduces testing overhead and hallucinations, and leads to more accurate and stable navigation.

Specifically, in Aux-Think, the generation of CoT reasoning and navigation actions is decoupled into two distinct tasks during training: (1) generating navigation actions based on stepwise observations and language instructions as the primary task, and (2) generating the reasoning process for each step as an auxiliary task. By leveraging prompts to switch between these tasks, we independently supervise navigation actions and reasoning processes, effectively avoiding the negative interference caused by jointly training both tasks. During testing, Aux-Think directly predicts actions without intermediate reasoning, thereby eliminating the risk of errors introduced by CoT hallucinations.

To validate the effectiveness of Aux-Think, we introduce R2R-CoT-320k, the first CoT dataset for VLN, which is  large-scale and specifically tailored for the R2R-CE benchmark \cite{krantz2020beyond}. 
As existing datasets lack aligned CoT-style reasoning, we construct this dataset by generating reasoning traces that can faithfully lead to the correct next action.
R2R-CoT-320k consists of over 320,000 diverse reasoning traces grounded in natural instructions and photo-realistic navigation trajectories. It covers a wide range of scenarios and CoT content, making it a rich and challenging resource for training and evaluation.
We show that Aux-Think, when trained with R2R-CoT-320k, matches the performance of state-of-the-art VLN methods, while using only a fraction of their training data (Fig. \ref{fig:1b}).

Our contributions are as follows: 
\begin{itemize}
    \item \textbf{New Finding:} We conduct a systematic comparison of reasoning strategies in VLN and reveal that test-time reasoning, including Pre-Think and Post-Think, consistently underperforms direct action prediction (No-Think), termed the TRC issue. To our knowledge, this is the first exploration of CoT strategies on the VLN-CE task.
    \item \textbf{New Method:} We propose Aux-Think, a novel training paradigm that uses CoT as auxiliary supervision while maintaining No-Think testing, achieving superior performance over other reasoning methods. Aux-Think pioneers a new perspective on CoT utilization and achieves the best performance on the navigation success rate.
    \item \textbf{New Dataset:}  We introduce R2R-CoT-320k, a large-scale, diverse, and challenging Chain-of-Thought dataset tailored for the R2R-CE benchmark, which enables more effective training of reasoning-aware VLN agents.
\end{itemize}

\begin{figure}[htbp]
    \centering
    \includegraphics[width=1\linewidth]{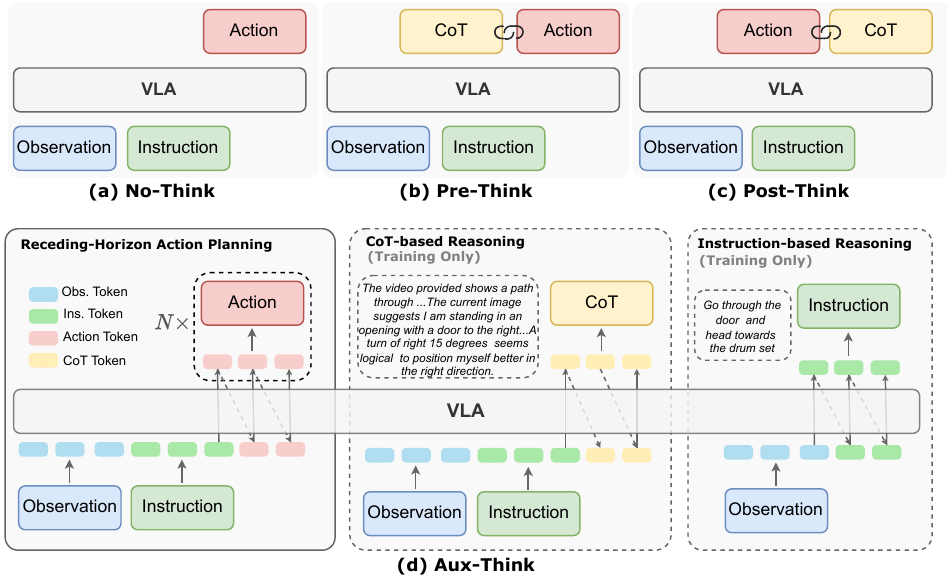}
    \caption{Illustration of Aux-Think and other reasoning strategies. Unlike No-Think, Pre-Think, and Post-Think, our Aux-Think introduces auxiliary CoT- and instruction-based reasoning during training while maintaining efficient action planning at testing.}
    \label{fig:pipeline}
\end{figure}

\section{Related Work}
\label{sec::related_work}
\subsection{Traditional VLN Methods}
Before the emergence of large-scale pretrained models, VLN methods are primarily built on  modular pipelines trained via imitation \cite{nguyen2019vision,wang2022towards,wu2020towards} or reinforcement learning \cite{xu2023benchmarking,wang2020vision}, often with handcrafted visual features and auxiliary objectives such as progress monitoring or instruction reweighting. These models typically operate with panoramic observations and discretized actions, as in benchmarks like Room-to-Room (R2R) \cite{anderson2018vision} and Room-Across-Room (RxR) \cite{ku2020room}. 
The traditional auxiliary reasoning tasks have also been explored, which rely on specific network outputs to predict structured, low-level reasoning results like task progress or trajectory alignment. 
More recent work has explored navigation in continuous action spaces using egocentric visual inputs \cite{krantz2020beyond,zhou2024navgpt}, which is the setting adopted in our study. Our method replaces handcrafted pipelines with vision-language models that directly predict agent actions, aiming to improve instruction following in realistic environments.

\subsection{VLN with Large Pretrained Models}

Recent advancements have seen the integration of large pre-trained models \cite{liu2024nvila, bai2025qwen2, yang2024qwen2, wang2025monodream}, into VLN tasks.
Early explorations of LLM in the VLN field usually use off-the-shelf large language models to select landmarks or waypoints in a zero-shot manner \cite{long2024discuss,long2024instructnav,shah2023lm,zhou2024navgpt}.
Recent works have focused on fine-tuning the VLM to obtain the navigational Vision-Language-Action model.
Notably, Poliformer \cite{zeng2024poliformer} and NaVid \cite{zhang2024navid} introduce a video-based monocular VLN, demonstrating navigation capabilities using monocular RGB videos without maps or depth input. Uni-NaVid \cite{zhang2024uni} unifies various navigation tasks, including VLN, ObjectNav \cite{chaplot2020object}, Embodied Question Answering \cite{das2018embodied}, and Human-following \cite{islam2019person,puig2023habitat}, into a single model trained on a diverse dataset. NaVILA \cite{cheng2024navila} further extends this approach by integrating VLN with legged robot locomotion skills in complex environments.

While these models have improved the alignment among visual understanding, language instructions, and navigation actions, they predominantly employ No-Think testing strategies, lacking reasoning mechanisms. Moreover, their performance gains often stem from leveraging extensive datasets, whereas our approach focuses on exploring reasoning strategies. % without increasing data requirements.

% \subsection{Reasoning in Large Vision-Language Models}
\subsection{Reasoning Models}

Recent advances like Chain-of-Thought (CoT) \cite{wei2022chain}, ReAct \cite{yao2023react}, and Toolformer \cite{schick2023toolformer} highlight the potential of LLMs to perform explicit reasoning in static and multimodal tasks, including VQA \cite{liu2025visual}, visual grounding \cite{liao2025improved}, and video understanding \cite{feng2025video}, where Pre-Think strategies have shown success. Similar ideas have been explored in embodied tasks like manipulation \cite{wen2024diffusion} and control \cite{zawalski2024robotic}. However, a recent study \cite{li2025think} argues that small models may benefit more from No-Think or Post-Think strategies due to limited CoT quality.

In our work, we conduct the first systematic comparison of Pre-Think, Post-Think, and No-Think reasoning strategies for VLN.
Based on our findings, we propose Aux-Think, a novel framework that leverages CoT reasoning as auxiliary supervision during training while maintaining No-Think testing, thereby enhancing data efficiency and performance in VLN.
% , deliberately avoiding RL-based methods to ensure a fair and controlled evaluation. 

\section{Method}
\label{sec::method}

\subsection{Problem Setup}

We study monocular Vision-and-Language Navigation in continuous environments (VLN-CE), where an embodied agent navigates photo-realistic indoor environments by following natural language instructions. VLN-CE emphasizes generalization to unseen environments and supports both forward and reverse navigation, offering a comprehensive test of spatial reasoning and language grounding.

At each time step, the agent receives: (1) a natural language instruction, typically a short paragraph specifying the navigation goal; (2) a RGB observation from the agent’s current viewpoint; and (3) historical observations, including 8 frames uniformly sampled from all historical frames (always including the first frame).
The agent selects an action (e.g., move forward, turn left/right by a specific degree, or stop). 
The objective is to generate an action sequence that follows the instruction as accurately and efficiently as possible until the agent reaches the target position.

Within the Supervised Fine-Tuning (SFT) framework, our VLN model based on NVILA 8B \cite{liu2024nvila}, is learned by imitating expert demonstrations from the dataset, where each trajectory provides sequences of <navigation context, expert action> pairs, with navigation context denoting the combination of historical observations, the current observation, and the instruction.

\subsection{R2R-CoT-320k Dataset Construction}
\label{sec:data}
\begin{figure}[htbp]
    \centering
    \includegraphics[width=0.95\linewidth]{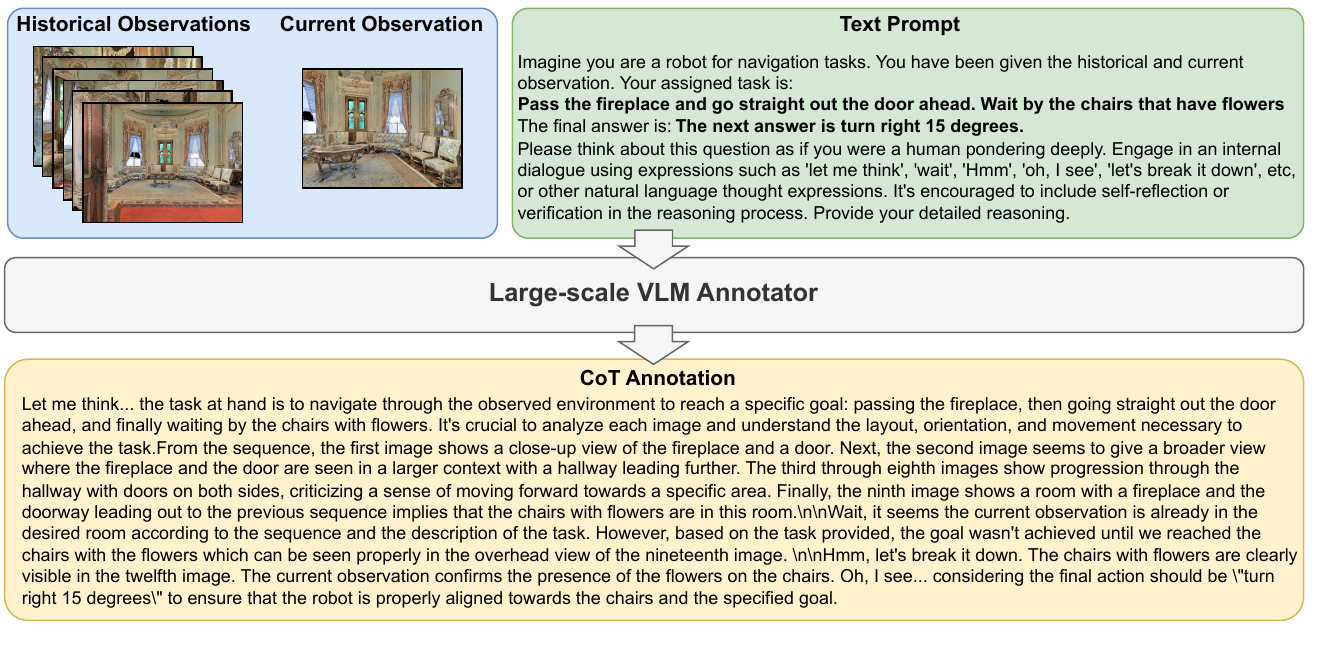}
    \caption{The annotation pipeline of our R2R-CoT-320k dataset.}
    \label{fig:label_visulization}
\end{figure}

We present R2R-CoT-320k, the first VLN dataset annotated with CoT reasoning, tailored for the R2R-CE benchmark.
We reconstruct step-wise navigation trajectories in the Habitat simulator \cite{habitat19iccv}. Each sample in the dataset comprises the current view, the historical visual context, the corresponding instruction, and the ground-truth action.
We employ Qwen-2.5-VL-72B \cite{bai2025qwen2}, one of the strongest publicly available VLMs, to generate detailed CoT for each navigation sample (Fig. \ref{fig:label_visulization}).
For Pre-Think and Post-Think strategies, we format reasoning traces with <think></think> and <answer></answer> tags, following recent reasoning models \cite{guo2025deepseek}. See Appendix \ref{sec:dataset} for more details on our dataset.

\subsection{Systematic Investigation on Reasoning Strategies for VLN}
\label{sec:reason}
To investigate the impact of reasoning on VLN, we study and evaluate three distinct strategies for integrating Chain-of-Thought (CoT) reasoning during training and testing.
\begin{wrapfigure}{r}{0.4\linewidth} % r=右侧，l=左侧
    \centering
    \includegraphics[width=\linewidth]{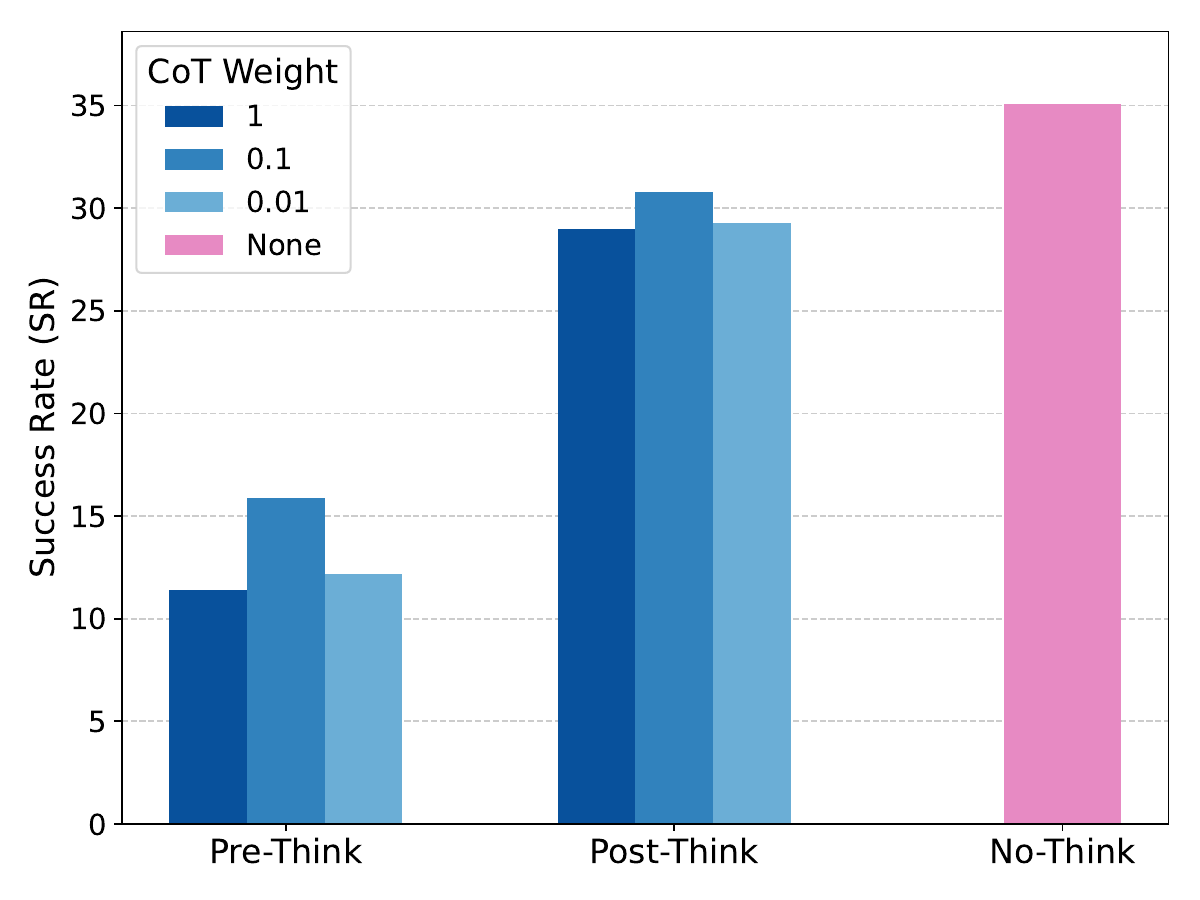}
    \caption{Comparision of success rate on Pre-Think, Post-Think, and No-Think.}
    \label{fig:think_sr}
    \vspace{-10mm}
\end{wrapfigure}
\textbf{No-Think:} The agent directly predicts the next action based on the current observation and instruction, without any intermediate reasoning. 

\textbf{Pre-Think:} The agent first generates an explicit reasoning trace based on the instruction and current observation. The following predicted actions are conditioned on the CoT output.

\textbf{Post-Think:} The agent first predicts an action and then retrospectively generates a reasoning trace that explains the decision.

The training loss for the VLN model $\pi_{\theta}$ with above three strategies is:
\begin{equation}
\label{equ:think_loss}
L(\theta) = - \sum_{\tau \in D} \sum_{t=0}^{T}
\begin{cases}
\log \pi_{\theta}(a_t^* | \mathcal{O}_t, o_t, I_\tau) & \text{for No-Think} \\
\log \pi_{\theta}(<c_t^*,a_t^*> | \mathcal{O}_t, o_t, I_\tau) & \text{for Pre-Think} \\
\log \pi_{\theta}(<a_t^*,c_t^*> | \mathcal{O}_t, o_t, I_\tau) & \text{for Post-Think}
\end{cases}
\end{equation}
where $D$ is the training dataset; $T$ is the total number of timesteps in a trajectory $\tau$, and $t$ is the current timestep. $a_t^*$ is the ground truth action and $c_t^*$ is the ground truth reasoning trace. $\mathcal{O}_t$ represents the history of observations up to timestep $t$, $o_t$ is the current visual observation at timestep $t$, and $I_\tau$ is the natural language instruction provided to the agent.

We isolate the impact of reasoning on overall navigation performance. In addition, we adjust the loss weight of the CoT part ($c_t^*$ in Equation \ref{equ:think_loss}) to make the model more focused on learning the action (Fig. \ref{fig:think_sr}). Our key findings include:

% \begin{figure}[htbp]
%     \centering 
%     \begin{subfigure}[t]{0.45\textwidth} 
%         \centering % Center content within the subfigure
%         \includegraphics[width=0.99\linewidth]{pic/think_model.pdf}
%         \caption{Illustration of different reasoning strategies.}
%         \label{fig:2a} % Changed label to reflect it's part (a)
%     \end{subfigure}
%     \hspace{0.05\textwidth} % Adjust spacing between subfigures if needed
%     % Subfigure (b)
%     \begin{subfigure}[t]{0.48\textwidth} % Or adjust width as needed, [t] aligns tops
%         \centering % Center content within the subfigure
%         \includegraphics[width=0.8\linewidth]{pic/think_sr.pdf}
%         \caption{Comparision of success rate on Pre-Think, Post-Think, and No-Think Strategies.} % Note: "method name" seems like a placeholder here
%         \label{fig:2b} % Added a label for part (b)
%     \end{subfigure}
%     \caption{Reasoning strategies and their performance. More detailed results are provided in the Section \ref{sec:exp_reason} and Appendix \ref{sec:more}.} % Add an overall caption for the figure
%     \label{fig:think}
% \end{figure}

% \begin{figure}[htbp]
%     \centering 
%         \includegraphics[width=0.4\linewidth]{pic/think_sr.pdf}
%         \caption{Comparision of success rate on Pre-Think, Post-Think, and No-Think Strategies.} % Note: "method name" seems like a placeholder here
%     \label{fig:think_sr}
% \end{figure}

% \vspace{0.5em}
\noindent
\begin{tcolorbox} %[colback=gray!10, colframe=gray!30, title=Findings]
\textbf{Finding 1: Pre-Think and Post-Think perform significantly worse than No-Think.} Despite involving explicit reasoning, both strategies lead to lower navigation success and efficiency, highlighting the unreliability of test-time CoT in dynamic environments.

\

\textbf{Finding 2: Careful Balancing of Explicit CoT Loss Weight Improves Performance.}  By carefully tuning the explicit CoT weight for Pre-Think and Post-Think, we find that balanced supervision yields slight gains, indicating that training emphasis on reasoning is a key, strategy-dependent factor despite test-time unreliability.
\end{tcolorbox}

\color{black}

We further analyze the severe performance degradation observed with Pre-Think and Post-Think strategies. 
During training, the model is only exposed to optimal sequences of states and actions in oracle trajectories.
However, VLN environments are inherently complex, dynamic, and partially observable, increasing the likelihood of agents deviating from optimal, oracle-guided training trajectories. Consequently, when encountering non-oracle, out-of-distribution states during testing, the CoT reasoning generated by these strategies is susceptible to drift, potentially yielding inaccurate or hallucinated interpretations of the environment and instructions.

In Pre-Think, the model’s actions rely heavily on the preceding reasoning chain, making decisions fragile. Any hallucination or misstep in reasoning directly leads to wrong actions.
In Post-Think, although actions are generated first, the need to produce follow-up explanations still alters the model’s hidden states. This subtle interference, like reserving capacity or shifting attention, can compromise the quality of the initial action decision. A detailed example is provided in Fig. \ref{fig:visulization}.

Moreover, agents perform multi-step action prediction, where an error at any intermediate step further pushes the agent away from the correct state distribution. This compounding effect leads to cascading errors in both CoT reasoning and subsequent actions, ultimately resulting in trajectory-level failures.

\subsection{Aux-Think: Reasoning-Aware Co-Training Strategy}
\label{sec:aux_think}

To address the challenges from the explicit CoT training to VLN, we propose Aux-Think, a novel reasoning strategy designed to enhance navigation performance without incurring test-time problems.
Aux-Think leverages reasoning exclusively during training through auxiliary tasks.
We design two reasoning-based auxiliary tasks and one action-based primary task during training.

\textbf{CoT-based Reasoning.} The model is trained to generate CoT traces conditioned on the given instruction $I$, current observation $o_t$, and historical observations $\mathcal{O}_t$. This encourages the acquisition of structured reasoning patterns and strengthens the connection between language, vision, and actions. The loss of CoT-based reasoning for each trajectory $\tau$ is:
\begin{equation}
    L^{CoT}_{\tau}(\theta) = - \sum_{t=0}^{T} \log \pi_{\theta}(c_t^* | \mathcal{O}_t, o_t, I_\tau)
\end{equation}
where $c_t^*$ is the ground-truth reasoning trace at step $t$.
    
\textbf{Instruction-based Reasoning.} Given a sequence of visual observations $\sum_{t=0}^T o_t$, the model is trained to reconstruct the corresponding instruction $I$. This reverse reasoning task provides complementary supervision beyond CoT-based signals, further enriching the model's semantic grounding. The training loss is:
\begin{equation}
    L^{Ins}_{\tau}(\theta) = - \log \pi_{\theta}(I_\tau | \sum_{t=0}^{T} o_t)
\end{equation}
% where  represents a temporally aggregated observed images of the trajectory.
\textbf{Receding-Horizon Action Planning.} We introduce Receding-Horizon Action Planning as our primary task. During training, the model predicts a sequence of the next $n$ actions $(a_t, a_{t+1},..., a_{t+n-1})$ based on the instruction $I$, current observation $o_t$, and navigation history $\mathcal{O}_t$ for the sample at time step $t$ . This setup encourages short-term forecasting while retaining reactivity to new observations. The training objective for each trajectory $\tau$ is defined as:
\begin{equation}
L^{Act}_{\tau}(\theta) = -\sum_{t=0}^{T} \sum_{k=0}^{n} \log \pi_{\theta}(a_{t+k}^* | \mathcal{O}_t, o_t, I_\tau)
\end{equation}
where $a_{t+k}^*$ denotes the ground-truth action at future step $t+k$.

During training, we co-train the three tasks and switch between different tasks by changing the prompt (Appendix \ref{sec:prompt}). The final loss function is:
\begin{equation}
    L = \sum_{\tau \in D} L^{Act}_{\tau}(\theta) + L^{CoT}_{\tau}(\theta) + L^{Ins}_{\tau}(\theta)
\end{equation}
where $D$ is the set of training trajectories.

For testing, we only activate the prediction of actions, and the model predicts the next $n$ actions and executes only the first one. This ensures fast, reactive navigation without reasoning overhead. 
We demonstrate its stabilizing effect in long-horizon trajectories through ablation studies (Table \ref{tab:action}).

\section{Experimental Results}
\label{sec::experimental}

\subsection{Experiment Setup}
\label{sec:exp_set}

\textbf{Simulated environments.} We evaluate our method on the VLN-CE benchmarks R2R-CE \cite{krantz2020beyond} and RxR-CE \cite{ku2020room} following the standard VLN-CE settings. 
All the methods are evaluated on the R2R val-unseen split and RxR val-unseen split.

\textbf{Metrics.} We follow the standard VLN evaluation protocol \cite{krantz2020beyond, ku2020room} to evaluate the navigation performance for all the methods, including success rate (SR), oracle success rate (OSR), success weighted by path length (SPL), and navigation error from goal (NE).

\subsection{Implementation Details} 
\label{sec:exp_implementation}

\textbf{Model training.} We use NVILA-lite 8B \cite{liu2024nvila} as the base pretrained model, which consists of a vision encoder (SigLIP \cite{tschannen2025siglip}), a projector, and an LLM (Qwen 2 \cite{bai2025qwen2}). We use supervised finetuning (SFT) to train our VLN model from stage 2 of NVILA-lite, as it has finished visual language corpus pre-training.
Our model is trained with 8 NVIDIA H20 GPUs for one epoch (around 60 hours), with a learning rate of 1e-5.
% In addition, we also try reinforcement fine-tuning, referring to Appendix \ref{sec:rl}.

\textbf{Action design.} 
The action space is designed into four categories: move forward, turn left, turn right, and stop. The forward action includes step sizes of 25 cm, 50 cm, and 75 cm, while the turn actions are parameterized by rotation angles of 15$^\circ$ , 30$^\circ$ , and 45$^\circ$ . This fine-grained design allows for more precise and flexible control, which is critical in complex environments.

\subsection{Comparison on VLN-CE Benchmarks}

We evaluate our method on the VLN-CE benchmarks, which provide continuous environments for navigational actions in reconstructed photorealistic indoor scenes.
We first focus on the val-unseen split in \textbf{R2R-CE} dataset in Table \ref{tab:r2r_val_unseen}.
To be fair, we distinguish between methods by marking those based on waypoint predictors ($\ast$) and those that are not based on large models ($\dagger$).

In large model-based methods, we additionally mark the amount of data used by the method in addition to the R2R-CE training split (\textbf{Extra Data}).
We further scale up by the RxR training split (600K), DAgger data (500K) and web data (500K), and our performance achieves the SOTA Success Rate (SR) to those using a much larger amount of training data.

\begin{table}[htbp]
  \centering
  \caption{Comparison of different methods on the R2R Val-Unseen split. Observations used include Monocular (Mono.) and Panoramic view (Pano.). $\ast$ indicates methods based on the waypoint predictor \cite{hong2022bridging}. $\dagger$ indicates methods without using LLMs. $^\circ$ indicates the models using the training data only from R2R-CE training split, and we compare with the results reported in their paper for a fair evaluation. The training data structures of traditional non-LLM-based methods are quite different, so we do not compare them with their extra data.}
  \label{tab:r2r_val_unseen}
  % \sisetup{detect-weight=true, detect-inline-weight=math} % Ensure bold numbers are handled correctly by S columns
  % Consider removing resizebox or using \small/\footnotesize for better readability
  \resizebox{0.9\textwidth}{!}{
  % \begin{tabular}{l c cc S[table-format=1.2] S[table-format=2.1, parse-numbers=false] S[table-format=2.1, parse-numbers=false] S[table-format=2.1, parse-numbers=false] r} 
  \begin{tabular}{l c cc c c c c r}
    \toprule
    % --- Header Row 1: Use multirow for Method & Venue headers ---
    \multirow{2}{*}{Method} & \multirow{2}{*}{Venue} & \multicolumn{2}{c}{Observation} & \multicolumn{4}{c}{R2R Val-Unseen} & {Training} \\ 
    \cmidrule(lr){3-4} \cmidrule(lr){5-8} \cmidrule(lr){9-9}
     & & {Mono.} & {Pano.} & {NE $\downarrow$} & {OSR $\uparrow$} & {SR $\uparrow$} & {SPL $\uparrow$} & {Extra Data} \\
    \midrule
    % --- Data Rows: Moved venue info to the second column ---
    BEVBert$^{\ast\dagger}$\cite{an2022bevbert}      & ICCV2023   &            & $\checkmark$  & 4.57 & 67.0 & 59.0 & 50.0 &   - \\
    ETPNav$^{\ast\dagger}$\cite{an2024etpnav}       & TPAMI2024  &            & $\checkmark$  & 4.71 & 65.0 & 57.0 & 49.0 &   - \\
    ENP–ETPNav$^{\ast\dagger}$\cite{liu2024vision} & Neurips2024 & & $\checkmark$  & 4.69 & {65}   & {58}   & {50}   &  - \\ % Added braces around numbers not matching S format if needed, or use parse-numbers=false
    \midrule
    Seq2Seq$^\dagger$\cite{krantz2020beyond}   & ECCV2020   & $\checkmark$  & & 7.77 & 37.0 & 25.0 & 22.0 &  -  \\
    CMA$^\dagger$\cite{krantz2020beyond}   & ECCV2020   & $\checkmark$  & & 7.37 & 40.0 & 32.0 & 30.0 &  -  \\
    LAW$^\dagger$\cite{raychaudhuri-etal-2021-language}  & EMNLP2021  & $\checkmark$  & & 6.83 & 44.0 & 35.0 & 31.0 &  -  \\
    CM2$^\dagger$\cite{georgakis2022cross} & CVPR2022   & $\checkmark$  & & 7.02 & 41.0 & 34.0 & 27.0 &  -  \\
    WS-MGMap$^\dagger$\cite{chen2022weakly} & Neurips2022 & $\checkmark$  & & 6.28 & 47.0 & 38.0 & 34.0 &   - \\
    sim2real$^\dagger$\cite{wang2024sim} & CoRL2024   & $\checkmark$  & & 5.95 & 55.8 & 44.9 & 30.4 &  -  \\
    \midrule
    % Open-Nav-Llama3.1\cite{qiao2024open}         & Arxiv2024  & $\checkmark$  & & 7.25 & {23}   & {16}   & 12.9 &   0K  \\
    NaVid$^\circ$\cite{zhang2024navid}         & RSS2024    & $\checkmark$  & & \underline{6.33} & \underline{30.8} & \underline{24.7} & \underline{23.6} &   0K  \\
    Aux-Think (ours)$^\circ$   & -          & $\checkmark$  & & \textbf{6.01} & \textbf{52.2} & \textbf{46.0} & \textbf{40.5} &  0K  \\ % Added '-' for missing venue
    \midrule
    Uni-NaVid\cite{zhang2024uni}         & RSS2025    & $\checkmark$  & & \underline{5.58} & 53.3 & 47.0 & 42.7 &  5570K  \\
    NaVILA\cite{cheng2024navila}   & RSS2025    & $\checkmark$  & & \textbf{5.22} & \textbf{62.5} & \underline{54.0} & \textbf{49.0} & 2770K   \\
    Aux-Think (ours)   & -          & $\checkmark$  & & 6.08 & \underline{60.0} & \textbf{54.8} & \underline{46.9} & \textbf{1600K}  \\ % Added '-' for missing venue
    \bottomrule
  \end{tabular}}
\end{table}

As in Table \ref{tab:r2r_val_unseen}, our proposed \textbf{Aux-Think} achieves strong performance with and without extra data.
We attribute these results to multilevel reasoning supervision. Our joint training on CoT-based reasoning, instruction reconstruction, and receding-horizon action prediction enriches the model’s semantic grounding and decision-making ability, allowing it to better generalize from limited data.
Our method benefits from reasoning-induced supervision signals that align more closely with the high-level semantic structure of the instructions, making each training example more informative.

To further assess the ability of Aux-Think, we evaluate it on the RxR-CE \cite{ku2020room} Val-Unseen split (Table~\ref{tab:app_rxr_val_unseen}). Compared to R2R-CE, RxR-CE includes more natural instructions and longer trajectories, making it a more realistic and challenging benchmark. Aux-Think achieves strong overall performance, particularly on the Success Rate (SR) metric, where it surpasses Uni-NaVid and NaVILA while using much fewer training data (1920K vs. 5900K and 3100K). This demonstrates the effectiveness of reasoning supervision under limited data.
It is noted that performance on NE and SPL is relatively modest. This is likely due to the model's internalized reasoning behavior, learned during CoT training, which encourages broader exploration. Although this can lead to longer paths and reduced efficiency, it improves SR and OSR by increasing the likelihood of reaching the goal, especially in unfamiliar environments.

\begin{table}[htbp]
  \centering
  \caption{Comparison of different methods on the RxR Val-Unseen split. $\dagger$ indicates methods without using LLMs. The training data structures of traditional non-LLM-based methods are quite different, so we do not include their training data in this table.}
  \label{tab:app_rxr_val_unseen}
  % \sisetup{detect-weight=true, detect-inline-weight=math} % Ensure bold numbers are handled correctly by S columns
  % Consider removing resizebox or using \small/\footnotesize for better readability
  \resizebox{0.9\textwidth}{!}{
  % \begin{tabular}{l c cc S[table-format=1.2] S[table-format=2.1, parse-numbers=false] S[table-format=2.1, parse-numbers=false] S[table-format=2.1, parse-numbers=false] r} 
  \begin{tabular}{l c cc c c c c r}
    \toprule
    % --- Header Row 1: Use multirow for Method & Venue headers ---
    \multirow{2}{*}{Method} & \multirow{2}{*}{Venue} & \multicolumn{2}{c}{Observation} & \multicolumn{4}{c}{RxR Val-Unseen} & {Training} \\ % Method/Venue span 2 rows. Obs/R2R span multiple cols. Training is super-header.
    % --- Header Row 2: Sub-headers and column rules ---
    \cmidrule(lr){3-4} \cmidrule(lr){5-8} \cmidrule(lr){9-9} % Adjust cmidrule ranges for new column layout
    % Headers for columns 3 to 9. Cols 1 & 2 are covered by \multirow above.
     & & {Mono.} & {Pano.} & {NE $\downarrow$} & {OSR $\uparrow$} & {SR $\uparrow$} & {SPL $\uparrow$} & { Data} \\
     \midrule
     % BEVBert$^{\dagger}$\cite{an2022bevbert}      & ICCV2023   &            & \checkmark & 4.00 & - & 68.5 & - &   - \\
     ETPNav$^{\dagger}$\cite{an2024etpnav}       & TPAMI2024  &            & \checkmark & 5.64 & - & 54.7 & 44.8 &   - \\
     ENP–ETPNav$^{\dagger}$\cite{liu2024vision} & Neurips2024 & & $\checkmark$ & 5.51 & -   & 55.27   &  45.11   &  - \\
    \midrule
    Seq2Seq$^\dagger$\cite{krantz2020beyond}   & ECCV2020 & $\checkmark$  &  & 11.8& - & 13.9 & 11.9 &  -  \\
    LAW$^\dagger$\cite{raychaudhuri-etal-2021-language}  & EMNLP2021  & $\checkmark$  & & 10.87 & 21.0 & 8.0 & 8.0 &  -  \\
    CM2$^\dagger$\cite{georgakis2022cross} & CVPR2022  & $\checkmark$  &  & 12.29 & 25.3 & 14.4 & 9.2 &  -  \\
    sim2real$^\dagger$\cite{wang2024sim} & CoRL2024  & $\checkmark$  &  & 8.79 & 36.7 & 25.5 & 18.1 &  -  \\
    \midrule
    Uni-NaVid\cite{zhang2024uni}         & RSS2025 & $\checkmark$  &    & \textbf{6.24} & \underline{55.5} & 48.7 & \underline{40.9} &  5900K  \\
    NaVILA\cite{cheng2024navila}   & RSS2025   & $\checkmark$  &  & \underline{6.77} & - & \underline{49.3} & \textbf{44.0} & 3100K   \\
    Aux-Think (ours)   & -    & $\checkmark$  &    & \textbf{6.24} & \textbf{61.9} & \textbf{52.2} & 40.2 & \textbf{1920K}  \\ % Added '-' for missing venue
    \bottomrule
  \end{tabular}}
\end{table}
\subsection{Comparison Between Different Reasoning Strategies}
\label{sec:exp_reason}

% , with  only 32.5\% and 82.6\% respectively. The Aux-Think we propose increases the SR from 35.1\% to 41.3\%.
As shown in Table \ref{tab:reason}, we compare different reasoning strategies on R2R-CE, with only the R2R-CoT-320K dataset as training data for fairness.
We find that the SR performance of Pre-Think and Post-Think is significantly lower than No-Think.
In Pre-Think, the action prediction is conditioned on the generated CoT; thus, low-quality or poorly learned CoT directly impairs action accuracy. While Post-Think partially mitigates this issue by generating CoT after the action, suboptimal CoT representations can still degrade overall performance. In contrast, the proposed Aux-Think decouples CoT and action learning by implicitly internalizing CoT knowledge into feature representations. 

\begin{figure}[htbp]
    \centering 
    \begin{subfigure}[t]{0.45\textwidth} 
        \centering
        \includegraphics[width=0.84\linewidth]{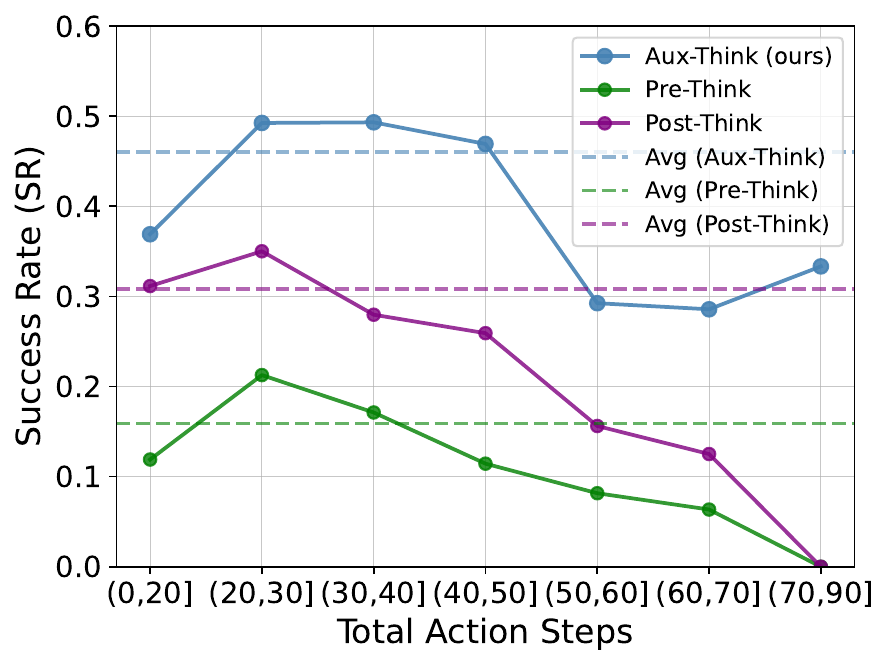}
        \caption{}
        \label{fig:step_sr}
    \end{subfigure}
    \hspace{0.05\textwidth} % Adjust spacing between subfigures if needed
    % Subfigure (b)
    \begin{subfigure}[t]{0.45\textwidth} 
        \centering
        \includegraphics[width=0.84\linewidth]{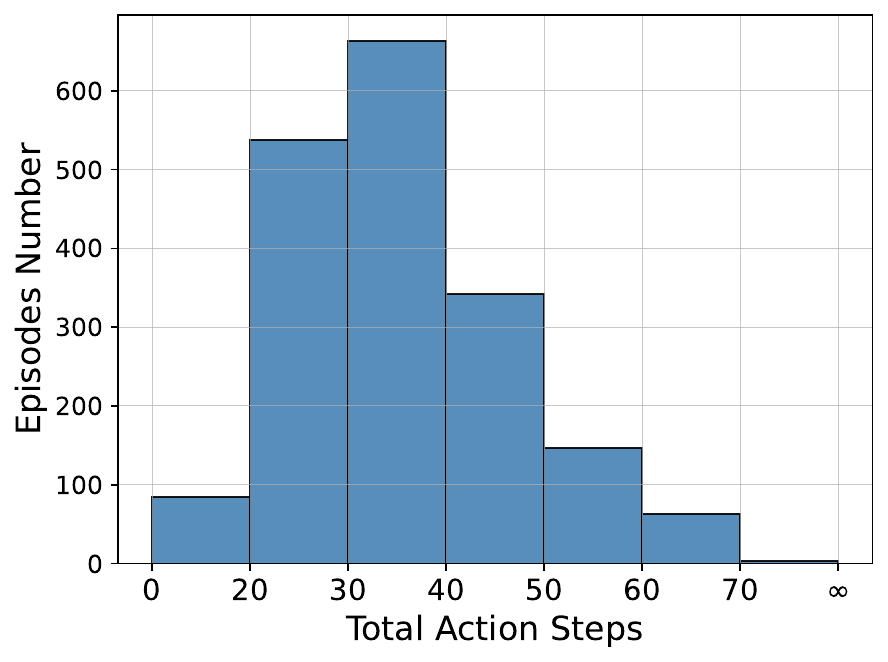}
        \caption{}
        \label{fig:train_step}
    \end{subfigure}
    \caption{(a) Success Rate of reasoning strategies on different steps during testing. (b) The epsodes number on different steps during testing. The steps indicate the actions required by the instruction.} % Add an overall caption for the figure
    \label{fig:think_step}
\end{figure}

% \vspace{-0.3cm}
\begin{table}[htbp]
\centering
\caption{Comparison of different reasoning strategies on R2R-CE Val-Unseen split.}
\label{tab:reason}
\scalebox{0.9}{
\begin{tabular}{c|cccc|c}
\toprule
Reason Strategies & \multicolumn{1}{c}{NE$\downarrow$}                         & \multicolumn{1}{c}{OSR$\uparrow$}                        & \multicolumn{1}{c}{SR$\uparrow$}                         & \multicolumn{1}{c}{SPL$\uparrow$}    & \multicolumn{1}{|l}{Avg. time$\downarrow$}                  \\ \midrule
No-Think          & 7.78 & 43.7 & 35.1 & 30.2  & \textbf{1.25s} \\
Pre-Think         & 9.23 & 19.3 & 11.4 & 8.6   & 30.62s     \\
Post-Think        & 8.59 & 35.1 & 29.0 & 23.8   & 28.97s    \\
Aux-Think (ours)  & \textbf{7.09} & \textbf{47.6}  & \textbf{41.3}   & \textbf{35.8} & \textbf{1.25s}                                \\ \bottomrule
\end{tabular}}
\end{table}
% \vspace{-0.3cm}
In Fig. \ref{fig:think_step}, we evaluate the Success Rate (SR) per test step for Aux-Think (ours), Pre-Think, and Post-Think, with results grouped by the number of steps required for task completion. Across all step ranges, Aux-Think consistently outperforms both baselines. A key observation is that the performance of Pre-Think and Post-Think degrades sharply as the required steps increase, with SR approaching zero for tasks exceeding 70 steps. In contrast, Aux-Think maintains strong performance even on longer-horizon tasks, exhibiting markedly higher robustness and generalization to complex, multi-step navigation scenarios. These results highlight the superior scalability of Aux-Think in handling extended reasoning and decision-making under increased task complexity.

\subsection{Ablation Studies}

\subsubsection{Impact of Different Auxiliary Tasks and Receding-Horizon Action Planning}

Table~\ref{tab:ablation} presents the ablation of three components: CoT-based Reasoning (A), Instruction-based Reasoning (B), and Receding-Horizon Action Planning (C). Introducing CoT Reasoning (A) leads to a noticeable improvement across all metrics, indicating its effectiveness in guiding action decisions. Adding Non-CoT Reasoning (A+B) further enhances performance, suggesting that the two forms of reasoning are complementary. The full model (A+B+C), which incorporates receding-horizon planning, achieves the best results, particularly in terms of SPL and SR, demonstrating that long-term planning grounded in implicit reasoning yields the most robust behavior. These results validate the necessity of integrating both reasoning and planning for optimal performance.

\vspace{-0.3cm}
\begin{table}[htbp]
\centering
\begin{minipage}[t]{0.48\linewidth}
\centering
\vspace{0pt}
\caption{Ablation study on different components.\\
A: CoT Reasoning, B: Non-CoT Reasoning, C: Receding-Horizon Action Planning.}
\scalebox{0.9}{
\begin{tabular}{ccc|cccc}
\toprule
\multicolumn{3}{c|}{Configuration} & \multicolumn{4}{c}{Metrics} \\
A & B & C & NE$\downarrow$ & OSR$\uparrow$ & SR$\uparrow$ & SPL$\uparrow$ \\ \midrule
  &   &   & 7.78 & 43.7 & 35.1 & 30.2 \\
\checkmark &   &   & 7.08 & 47.6 & 41.3 & 35.8 \\
 & \checkmark  &   & 7.12 & 46.3 & 40.6 & 35.7 \\
 &   & \checkmark  & 7.14 & 47.3 & 37.1 & 32.2 \\
\checkmark & \checkmark &   & 6.92 & 49.1 & 44.2 & 38.9 \\
\checkmark & \checkmark & \checkmark & \textbf{6.01} & \textbf{52.2} & \textbf{46.0} & \textbf{40.5} \\
\bottomrule
\end{tabular}}
\label{tab:ablation}
\end{minipage}
\hfill
\begin{minipage}[t]{0.48\linewidth}
\centering
\vspace{0pt}
\caption{Ablation studies on the predicted steps\\in our Receding-Horizon Action Planning. The best one across all metrics is when the number of steps is 3. }
\scalebox{0.98}{
\begin{tabular}{c|cccc}
\toprule
\#Steps & NE$\downarrow$ & OSR$\uparrow$ & SR$\uparrow$ & SPL$\uparrow$ \\ \midrule
1 & 7.78 & 43.7 & 35.1 & 30.2 \\
2 & 7.88 & 44.2 & 35.8 & 30.9 \\
3 & \textbf{7.14} & \textbf{47.3} & \textbf{41.4} & \textbf{36.1} \\
4 & 7.50 & 43.6 & 36.4 & 31.8 \\
5 & 7.54 & 43.7 & 37.1 & 32.2 \\
\bottomrule
\end{tabular}}
\label{tab:action}
\end{minipage}
\end{table}

% \vspace{-0.3cm}
\subsubsection{Impact of Steps in Receding-Horizon Action Planning}

Based on Table \ref{tab:action}, the model achieves the best performance when the number of predicted steps is set to 3. 
We disable the CoT auxiliary task to analyze the impact of actions more clearly.
The results highlight our Receding-Horizon Action Planning, which encourages the model to anticipate future actions and enhances its planning capabilities. However, increasing the number of predicted steps beyond this point leads to performance degradation. We attribute this to the limited perceptual field of monocular observations without additional global knowledge, which makes long-horizon prediction more challenging and can cause the model to generate suboptimal or collapsed navigation behaviors.

\section{Limitation and Future Work}
\label{sec:limit}

This work evaluates Aux-Think’s data efficiency under a controlled, widely adopted setup: SFT on the R2R dataset with monocular RGB input. This enables fair comparison and isolates the effect of reasoning-aware supervision.
While constrained, this setting opens future directions, scaling to larger navigation datasets and incorporating richer supervision (e.g., depth, panorama, localization) \cite{wang2025monodream}.
% However, we have not found an effective way to to improve both CoT and action quality through reinforcement learning jointly, and exploring reinforcement learning with lightweight VLMs like SmolVLM2 \cite{marafioti2025smolvlm} for more scalable policy learning  will be our future work.

\section{Conclusion}
We conduct the first systematic investigation of reasoning strategies in Vision-and-Language Navigation, revealing a key limitation, \textit{Test-time Reasoning Collapse}, where errors in generated reasoning can compound and degrade navigation performance. Motivated by this finding, we propose \textbf{Aux-Think}, a reasoning-aware co-training framework that leverages Chain-of-Thought as auxiliary supervision during training, while relying on efficient No-Think testing. Extensive experiments demonstrate that Aux-Think achieves performance on par with state-of-the-art methods while using significantly less training data, highlighting its robustness and data efficiency. We also release \textbf{R2R-CoT-320k}, the first CoT dataset for VLN, to facilitate future research on reasoning models.

\section{Acknowledgement}
This work was supported by the National Natural Science Foundation of China under Grant No. 62441617. It was supported by the Postdoctoral Fellowship Program and China Postdoctoral Science Foundation under Grant No. 2024M764093 and Grant No. BX20250485, the Beijing Natural Science Foundation under Grant No. 4254100, the Fundamental Research Funds for the Central Universities under Grant No. KG16336301, and by Beijing Advanced Innovation Center for Future Blockchain and Privacy Computing.
Deying Li was supported in part by the National Natural Science Foundation of China Grant No.12071478. Yongcai Wang was supported in part by the National Natural Science Foundation of China Grant No. 61972404, Public Computing Cloud, Renmin University of China, and the Blockchain Lab, School of Information, Renmin University of China.
Shuo Wang was supported in part by the Outstanding Innovative Talents Cultivation Funded Programs 2024 of Renmin University of China.

\bibliographystyle{unsrt}
\bibliography{reference}  % .bib

\newpage
\section*{NeurIPS Paper Checklist}

\begin{enumerate}

\item {\bf Claims}
    \item[] Question: Do the main claims made in the abstract and introduction accurately reflect the paper's contributions and scope?
    \item[] Answer: \answerYes{} % Replace by \answerYes{}, \answerNo{}, or \answerNA{}.
    \item[] Justification: Refer to Abstract and Section \ref{sec:intro}.
    \item[] Guidelines:
    \begin{itemize}
        \item The answer NA means that the abstract and introduction do not include the claims made in the paper.
        \item The abstract and/or introduction should clearly state the claims made, including the contributions made in the paper and important assumptions and limitations. A No or NA answer to this question will not be perceived well by the reviewers. 
        \item The claims made should match theoretical and experimental results, and reflect how much the results can be expected to generalize to other settings. 
        \item It is fine to include aspirational goals as motivation as long as it is clear that these goals are not attained by the paper. 
    \end{itemize}

\item {\bf Limitations}
    \item[] Question: Does the paper discuss the limitations of the work performed by the authors?
    \item[] Answer: \answerYes{} % Replace by \answerYes{}, \answerNo{}, or \answerNA{}.
    \item[] Justification: Refer to Section \ref{sec:limit}.
    \item[] Guidelines:
    \begin{itemize}
        \item The answer NA means that the paper has no limitation while the answer No means that the paper has limitations, but those are not discussed in the paper. 
        \item The authors are encouraged to create a separate "Limitations" section in their paper.
        \item The paper should point out any strong assumptions and how robust the results are to violations of these assumptions (e.g., independence assumptions, noiseless settings, model well-specification, asymptotic approximations only holding locally). The authors should reflect on how these assumptions might be violated in practice and what the implications would be.
        \item The authors should reflect on the scope of the claims made, e.g., if the approach was only tested on a few datasets or with a few runs. In general, empirical results often depend on implicit assumptions, which should be articulated.
        \item The authors should reflect on the factors that influence the performance of the approach. For example, a facial recognition algorithm may perform poorly when image resolution is low or images are taken in low lighting. Or a speech-to-text system might not be used reliably to provide closed captions for online lectures because it fails to handle technical jargon.
        \item The authors should discuss the computational efficiency of the proposed algorithms and how they scale with dataset size.
        \item If applicable, the authors should discuss possible limitations of their approach to address problems of privacy and fairness.
        \item While the authors might fear that complete honesty about limitations might be used by reviewers as grounds for rejection, a worse outcome might be that reviewers discover limitations that aren't acknowledged in the paper. The authors should use their best judgment and recognize that individual actions in favor of transparency play an important role in developing norms that preserve the integrity of the community. Reviewers will be specifically instructed to not penalize honesty concerning limitations.
    \end{itemize}

\item {\bf Theory assumptions and proofs}
    \item[] Question: For each theoretical result, does the paper provide the full set of assumptions and a complete (and correct) proof?
    \item[] Answer: \answerYes{} % Replace by \answerYes{}, \answerNo{}, or \answerNA{}.
    \item[] Justification: Refer to Section \ref{sec:reason} and \ref{sec:aux_think}.
    \item[] Guidelines:
    \begin{itemize}
        \item The answer NA means that the paper does not include theoretical results. 
        \item All the theorems, formulas, and proofs in the paper should be numbered and cross-referenced.
        \item All assumptions should be clearly stated or referenced in the statement of any theorems.
        \item The proofs can either appear in the main paper or the supplemental material, but if they appear in the supplemental material, the authors are encouraged to provide a short proof sketch to provide intuition. 
        \item Inversely, any informal proof provided in the core of the paper should be complemented by formal proofs provided in appendix or supplemental material.
        \item Theorems and Lemmas that the proof relies upon should be properly referenced. 
    \end{itemize}

    \item {\bf Experimental result reproducibility}
    \item[] Question: Does the paper fully disclose all the information needed to reproduce the main experimental results of the paper to the extent that it affects the main claims and/or conclusions of the paper (regardless of whether the code and data are provided or not)?
    \item[] Answer: \answerYes{} % Replace by \answerYes{}, \answerNo{}, or \answerNA{}.
    \item[] Justification: Refer to Section \ref{sec:data} and \ref{sec:exp_set}
    \item[] Guidelines:
    \begin{itemize}
        \item The answer NA means that the paper does not include experiments.
        \item If the paper includes experiments, a No answer to this question will not be perceived well by the reviewers: Making the paper reproducible is important, regardless of whether the code and data are provided or not.
        \item If the contribution is a dataset and/or model, the authors should describe the steps taken to make their results reproducible or verifiable. 
        \item Depending on the contribution, reproducibility can be accomplished in various ways. For example, if the contribution is a novel architecture, describing the architecture fully might suffice, or if the contribution is a specific model and empirical evaluation, it may be necessary to either make it possible for others to replicate the model with the same dataset, or provide access to the model. In general. releasing code and data is often one good way to accomplish this, but reproducibility can also be provided via detailed instructions for how to replicate the results, access to a hosted model (e.g., in the case of a large language model), releasing of a model checkpoint, or other means that are appropriate to the research performed.
        \item While NeurIPS does not require releasing code, the conference does require all submissions to provide some reasonable avenue for reproducibility, which may depend on the nature of the contribution. For example
        \begin{enumerate}
            \item If the contribution is primarily a new algorithm, the paper should make it clear how to reproduce that algorithm.
            \item If the contribution is primarily a new model architecture, the paper should describe the architecture clearly and fully.
            \item If the contribution is a new model (e.g., a large language model), then there should either be a way to access this model for reproducing the results or a way to reproduce the model (e.g., with an open-source dataset or instructions for how to construct the dataset).
            \item We recognize that reproducibility may be tricky in some cases, in which case authors are welcome to describe the particular way they provide for reproducibility. In the case of closed-source models, it may be that access to the model is limited in some way (e.g., to registered users), but it should be possible for other researchers to have some path to reproducing or verifying the results.
        \end{enumerate}
    \end{itemize}

\item {\bf Open access to data and code}
    \item[] Question: Does the paper provide open access to the data and code, with sufficient instructions to faithfully reproduce the main experimental results, as described in supplemental material?
    \item[] Answer: \answerYes{} % Replace by \answerYes{}, \answerNo{}, or \answerNA{}.
    \item[] Justification: Refer to Section \ref{sec:data} and Appendix \ref{sec:dataset}
    \item[] Guidelines:
    \begin{itemize}
        \item The answer NA means that paper does not include experiments requiring code.
        \item Please see the NeurIPS code and data submission guidelines (\url{https://nips.cc/public/guides/CodeSubmissionPolicy}) for more details.
        \item While we encourage the release of code and data, we understand that this might not be possible, so “No” is an acceptable answer. Papers cannot be rejected simply for not including code, unless this is central to the contribution (e.g., for a new open-source benchmark).
        \item The instructions should contain the exact command and environment needed to run to reproduce the results. See the NeurIPS code and data submission guidelines (\url{https://nips.cc/public/guides/CodeSubmissionPolicy}) for more details.
        \item The authors should provide instructions on data access and preparation, including how to access the raw data, preprocessed data, intermediate data, and generated data, etc.
        \item The authors should provide scripts to reproduce all experimental results for the new proposed method and baselines. If only a subset of experiments are reproducible, they should state which ones are omitted from the script and why.
        \item At submission time, to preserve anonymity, the authors should release anonymized versions (if applicable).
        \item Providing as much information as possible in supplemental material (appended to the paper) is recommended, but including URLs to data and code is permitted.
    \end{itemize}

\item {\bf Experimental setting/details}
    \item[] Question: Does the paper specify all the training and test details (e.g., data splits, hyperparameters, how they were chosen, type of optimizer, etc.) necessary to understand the results?
    \item[] Answer: \answerYes{} % Replace by \answerYes{}, \answerNo{}, or \answerNA{}.
    \item[] Justification: Refer to Section \ref{sec::experimental}
    \item[] Guidelines:
    \begin{itemize}
        \item The answer NA means that the paper does not include experiments.
        \item The experimental setting should be presented in the core of the paper to a level of detail that is necessary to appreciate the results and make sense of them.
        \item The full details can be provided either with the code, in appendix, or as supplemental material.
    \end{itemize}

\item {\bf Experiment statistical significance}
    \item[] Question: Does the paper report error bars suitably and correctly defined or other appropriate information about the statistical significance of the experiments?
    \item[] Answer: \answerYes{} % Replace by \answerYes{}, \answerNo{}, or \answerNA{}.
    \item[] Justification: Refer to Secton \ref{sec::experimental}.
    \item[] Guidelines:
    \begin{itemize}
        \item The answer NA means that the paper does not include experiments.
        \item The authors should answer "Yes" if the results are accompanied by error bars, confidence intervals, or statistical significance tests, at least for the experiments that support the main claims of the paper.
        \item The factors of variability that the error bars are capturing should be clearly stated (for example, train/test split, initialization, random drawing of some parameter, or overall run with given experimental conditions).
        \item The method for calculating the error bars should be explained (closed form formula, call to a library function, bootstrap, etc.)
        \item The assumptions made should be given (e.g., Normally distributed errors).
        \item It should be clear whether the error bar is the standard deviation or the standard error of the mean.
        \item It is OK to report 1-sigma error bars, but one should state it. The authors should preferably report a 2-sigma error bar than state that they have a 96\% CI, if the hypothesis of Normality of errors is not verified.
        \item For asymmetric distributions, the authors should be careful not to show in tables or figures symmetric error bars that would yield results that are out of range (e.g. negative error rates).
        \item If error bars are reported in tables or plots, The authors should explain in the text how they were calculated and reference the corresponding figures or tables in the text.
    \end{itemize}

\item {\bf Experiments compute resources}
    \item[] Question: For each experiment, does the paper provide sufficient information on the computer resources (type of compute workers, memory, time of execution) needed to reproduce the experiments?
    \item[] Answer: \answerYes{} % Replace by \answerYes{}, \answerNo{}, or \answerNA{}.
    \item[] Justification: Refer to \ref{sec:exp_implementation}.
    \item[] Guidelines:
    \begin{itemize}
        \item The answer NA means that the paper does not include experiments.
        \item The paper should indicate the type of compute workers CPU or GPU, internal cluster, or cloud provider, including relevant memory and storage.
        \item The paper should provide the amount of compute required for each of the individual experimental runs as well as estimate the total compute. 
        \item The paper should disclose whether the full research project required more compute than the experiments reported in the paper (e.g., preliminary or failed experiments that didn't make it into the paper). 
    \end{itemize}
    
\item {\bf Code of ethics}
    \item[] Question: Does the research conducted in the paper conform, in every respect, with the NeurIPS Code of Ethics \url{https://neurips.cc/public/EthicsGuidelines}?
    \item[] Answer: \answerYes{} % Replace by \answerYes{}, \answerNo{}, or \answerNA{}.
    \item[] Justification: Refer to Appendix.
    \item[] Guidelines:
    \begin{itemize}
        \item The answer NA means that the authors have not reviewed the NeurIPS Code of Ethics.
        \item If the authors answer No, they should explain the special circumstances that require a deviation from the Code of Ethics.
        \item The authors should make sure to preserve anonymity (e.g., if there is a special consideration due to laws or regulations in their jurisdiction).
    \end{itemize}

\item {\bf Broader impacts}
    \item[] Question: Does the paper discuss both potential positive societal impacts and negative societal impacts of the work performed?
    \item[] Answer: \answerNA{} % Replace by \answerYes{}, \answerNo{}, or \answerNA{}.
    \item[] Justification: VLN research typically focuses on specific technical aspects or objectives that may not directly address broader societal impacts.
    \item[] Guidelines:
    \begin{itemize}
        \item The answer NA means that there is no societal impact of the work performed.
        \item If the authors answer NA or No, they should explain why their work has no societal impact or why the paper does not address societal impact.
        \item Examples of negative societal impacts include potential malicious or unintended uses (e.g., disinformation, generating fake profiles, surveillance), fairness considerations (e.g., deployment of technologies that could make decisions that unfairly impact specific groups), privacy considerations, and security considerations.
        \item The conference expects that many papers will be foundational research and not tied to particular applications, let alone deployments. However, if there is a direct path to any negative applications, the authors should point it out. For example, it is legitimate to point out that an improvement in the quality of generative models could be used to generate deepfakes for disinformation. On the other hand, it is not needed to point out that a generic algorithm for optimizing neural networks could enable people to train models that generate Deepfakes faster.
        \item The authors should consider possible harms that could arise when the technology is being used as intended and functioning correctly, harms that could arise when the technology is being used as intended but gives incorrect results, and harms following from (intentional or unintentional) misuse of the technology.
        \item If there are negative societal impacts, the authors could also discuss possible mitigation strategies (e.g., gated release of models, providing defenses in addition to attacks, mechanisms for monitoring misuse, mechanisms to monitor how a system learns from feedback over time, improving the efficiency and accessibility of ML).
    \end{itemize}
    
\item {\bf Safeguards}
    \item[] Question: Does the paper describe safeguards that have been put in place for responsible release of data or models that have a high risk for misuse (e.g., pretrained language models, image generators, or scraped datasets)?
    \item[] Answer: \answerNA{} % Replace by \answerYes{}, \answerNo{}, or \answerNA{}.
    \item[] Justification: VLN poses no such risks
    \item[] Guidelines:
    \begin{itemize}
        \item The answer NA means that the paper poses no such risks.
        \item Released models that have a high risk for misuse or dual-use should be released with necessary safeguards to allow for controlled use of the model, for example by requiring that users adhere to usage guidelines or restrictions to access the model or implementing safety filters. 
        \item Datasets that have been scraped from the Internet could pose safety risks. The authors should describe how they avoided releasing unsafe images.
        \item We recognize that providing effective safeguards is challenging, and many papers do not require this, but we encourage authors to take this into account and make a best faith effort.
    \end{itemize}

\item {\bf Licenses for existing assets}
    \item[] Question: Are the creators or original owners of assets (e.g., code, data, models), used in the paper, properly credited and are the license and terms of use explicitly mentioned and properly respected?
    \item[] Answer: \answerYes{} % Replace by \answerYes{}, \answerNo{}, or \answerNA{}.
    \item[] Justification: We have cited the corresponding original papers.
    \item[] Guidelines:
    \begin{itemize}
        \item The answer NA means that the paper does not use existing assets.
        \item The authors should cite the original paper that produced the code package or dataset.
        \item The authors should state which version of the asset is used and, if possible, include a URL.
        \item The name of the license (e.g., CC-BY 4.0) should be included for each asset.
        \item For scraped data from a particular source (e.g., website), the copyright and terms of service of that source should be provided.
        \item If assets are released, the license, copyright information, and terms of use in the package should be provided. For popular datasets, \url{paperswithcode.com/datasets} has curated licenses for some datasets. Their licensing guide can help determine the license of a dataset.
        \item For existing datasets that are re-packaged, both the original license and the license of the derived asset (if it has changed) should be provided.
        \item If this information is not available online, the authors are encouraged to reach out to the asset's creators.
    \end{itemize}

\item {\bf New assets}
    \item[] Question: Are new assets introduced in the paper well documented and is the documentation provided alongside the assets?
    \item[] Answer: \answerYes{} % Replace by \answerYes{}, \answerNo{}, or \answerNA{}.
    \item[] Justification: Refer to Section \ref{sec:data} and \ref{sec::experimental}.
    \item[] Guidelines:
    \begin{itemize}
        \item The answer NA means that the paper does not release new assets.
        \item Researchers should communicate the details of the dataset/code/model as part of their submissions via structured templates. This includes details about training, license, limitations, etc. 
        \item The paper should discuss whether and how consent was obtained from people whose asset is used.
        \item At submission time, remember to anonymize your assets (if applicable). You can either create an anonymized URL or include an anonymized zip file.
    \end{itemize}

\item {\bf Crowdsourcing and research with human subjects}
    \item[] Question: For crowdsourcing experiments and research with human subjects, does the paper include the full text of instructions given to participants and screenshots, if applicable, as well as details about compensation (if any)? 
    \item[] Answer: \answerNA{} % Replace by \answerYes{}, \answerNo{}, or \answerNA{}.
    \item[] Justification: This paper does not involve crowdsourcing nor research with human subjects.
    \item[] Guidelines:
    \begin{itemize}
        \item The answer NA means that the paper does not involve crowdsourcing nor research with human subjects.
        \item Including this information in the supplemental material is fine, but if the main contribution of the paper involves human subjects, then as much detail as possible should be included in the main paper. 
        \item According to the NeurIPS Code of Ethics, workers involved in data collection, curation, or other labor should be paid at least the minimum wage in the country of the data collector. 
    \end{itemize}

\item {\bf Institutional review board (IRB) approvals or equivalent for research with human subjects}
    \item[] Question: Does the paper describe potential risks incurred by study participants, whether such risks were disclosed to the subjects, and whether Institutional Review Board (IRB) approvals (or an equivalent approval/review based on the requirements of your country or institution) were obtained?
    \item[] Answer: \answerNA{}% Replace by \answerYes{}, \answerNo{}, or \answerNA{}.
    \item[] Justification: This paper does not involve crowdsourcing nor research with human subjects.
    \item[] Guidelines:
    \begin{itemize}
        \item The answer NA means that the paper does not involve crowdsourcing nor research with human subjects.
        \item Depending on the country in which research is conducted, IRB approval (or equivalent) may be required for any human subjects research. If you obtained IRB approval, you should clearly state this in the paper. 
        \item We recognize that the procedures for this may vary significantly between institutions and locations, and we expect authors to adhere to the NeurIPS Code of Ethics and the guidelines for their institution. 
        \item For initial submissions, do not include any information that would break anonymity (if applicable), such as the institution conducting the review.
    \end{itemize}

\item {\bf Declaration of LLM usage}
    \item[] Question: Does the paper describe the usage of LLMs if it is an important, original, or non-standard component of the core methods in this research? Note that if the LLM is used only for writing, editing, or formatting purposes and does not impact the core methodology, scientific rigorousness, or originality of the research, declaration is not required.
    %this research? 
    \item[] Answer: \answerNA{} % Replace by \answerYes{}, \answerNo{}, or \answerNA{}.
    \item[] Justification: Qwen 2.5 VL 72B was used to assist with CoT data annotation. However, the model was used strictly as a labeling aid under human supervision, and its outputs did not constitute a core, original, or non-standard component of the method itself. The annotated data was manually reviewed and curated, and the LLM's role was limited to speeding up the annotation process.
    \item[] Guidelines:
    \begin{itemize}
        \item The answer NA means that the core method development in this research does not involve LLMs as any important, original, or non-standard components.
        \item Please refer to our LLM policy (\url{https://neurips.cc/Conferences/2025/LLM}) for what should or should not be described.
    \end{itemize}

\end{enumerate}

%%%%%%%%%%%%%%%%%%%%%%%%%%%%%%%%%%%%%%%%%%%%%%%%%%%%%%%%%%%%
\newpage
\appendix

\section{Technical Appendices and Supplementary Material}

\subsection{More Experimental Results}
\label{sec:app_exp}

We evaluate the cross-data performance on \textbf{RxR-CE} Val-Unseen split, as shown in Table \ref{tab:rxr_results}. Even without using RxR-CE training data, our Aux-Think model achieves new state-of-the-art performance on the RxR-CE Val-Unseen split. This confirms the strong generalization of our reasoning-augmented co-training, enabling the model to transfer across datasets with different instructions and scenes.
\color{black}

\begin{table}[htbp]
\centering
\caption{Cross-dataset performance on the RxR-CE Val-Unseen split. All results are obtained without training on the RxR-CE training set. }
\label{tab:rxr_results}
\scalebox{0.9}{
\begin{tabular}{@{}l c c c r r r r@{}} % 8 columns: Method(l), Venue(l), Mono(c), Pano(c), NE(r), OSR(r), SR(r), SPL(r)
\toprule
\multirow{2}{*}{Method} & \multirow{2}{*}{Venue} & \multicolumn{2}{c}{Observation} & \multicolumn{4}{c}{RxR Val-Unseen} \\
\cmidrule(lr){3-4} \cmidrule(lr){5-8}
 & & {Mono.} & {Pano.} & {NE $\downarrow$} & {OSR $\uparrow$} & {SR $\uparrow$} & {SPL $\uparrow$} \\
\midrule
Seq2Seq\cite{krantz2020beyond}    & ECCV2020    & $\checkmark$ &   & 11.8            & 5.02            & 3.51            & 3.43            \\
CMA\cite{krantz2020beyond}       & ECCV2020    & $\checkmark$ &   & 11.7            & 10.7            & 4.41            & 2.47            \\
LAW\cite{raychaudhuri-etal-2021-language}  & EMNLP2021   & $\checkmark$ &   & 10.87           & 21.0            & 8.0             & 8.0             \\
CM2\cite{georgakis2022cross}     & CVPR2022    & $\checkmark$ &   & 8.98            & 25.3            & 14.4            & 9.2             \\
WS-MGMap\cite{chen2022weakly}    & Neurips2022 & $\checkmark$ &   & 9.83            & 29.8            & 15.0            & 12.1            \\
A$^2$NAV\cite{chen20232}    & Arxiv2023   & $\checkmark$ &   & {-}             & {-}             & 16.8            & 6.3             \\ % {-} is fine for 'r' column
NaVid\cite{zhang2024navid}       & RSS2024     & $\checkmark$ &   & \textbf{8.41}   & \underline{34.5}  & \underline{23.8}  & \underline{21.2}  \\
Aux-Think (ours)               & -           & $\checkmark$ &   & \underline{8.98}  & \textbf{39.6}   & \textbf{29.5}   & \textbf{23.6}   \\ % Added '-' for missing venue
\bottomrule
\end{tabular}}
\end{table}

\subsection{R2R-CoT-320k}
\label{sec:dataset}
The action labels in R2R-CoT-320k are derived from the original annotations in R2R-CE. To generate the Chain-of-Thought (CoT) annotations, we employ Qwen-VL 2.5 (72B). Specifically, for each navigation step, we provide the model with the agent's historical observations, the current visual input, and the next action. The model is then prompted to produce intermediate reasoning steps that reflect human-like decision-making processes. The annoation prompt is:

\begin{tcolorbox}[
  colback=gray!15,    % 背景颜色 (e.g., gray!10 for 10% gray, lightgray)
  colframe=black!75,  % 边框颜色 (e.g., black, darkgray)
  boxrule=0.5pt,      % 边框线条粗细
  arc=1mm,            % 边框圆角半径 (0mm for sharp corners)
  % title=Prompt Example, % 如果需要标题，取消注释这一行
]
Imagine you are a robot programmed for navigation tasks. You have been given a video of historical observations: <image>,...,<image> and and current observation: <image>. Your assigned task is: [Instruction]. Analyze this series of images to decide your next move, which could involve turning left or right by a specific degree, moving forward a certain distance, or stop if the task is completed.
The final answer is [Action].
Please think about this question as if you were a human pondering deeply. 
Engage in an internal dialogue using expressions such as 'let me think', 'wait', 'Hmm', 'oh, I see', 'let's break it down', etc, or other natural language thought expressions. It's encouraged to include self-reflection or verification in the reasoning process.
\end{tcolorbox}

\begin{figure}[htbp]
    \centering 
    \begin{subfigure}[t]{0.45\textwidth} 
        \centering
        \includegraphics[width=0.99\linewidth]{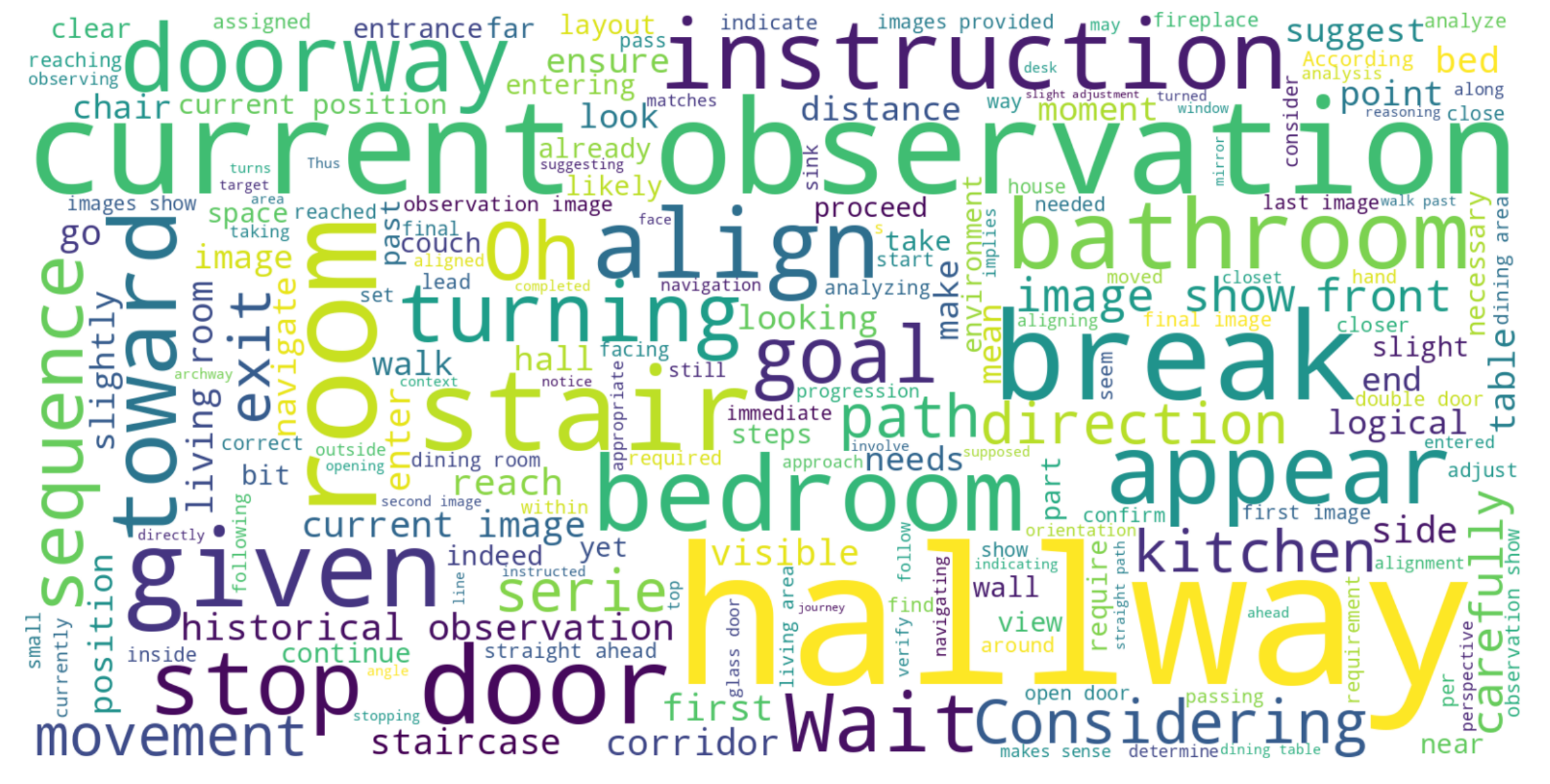}
        \caption{}
        \label{fig:word_cloud}
    \end{subfigure}
    % \hspace{0.05\textwidth} % Adjust spacing between subfigures if needed
    % Subfigure (b)
    \begin{subfigure}[t]{0.45\textwidth} 
        \centering
        \includegraphics[width=0.99\linewidth]{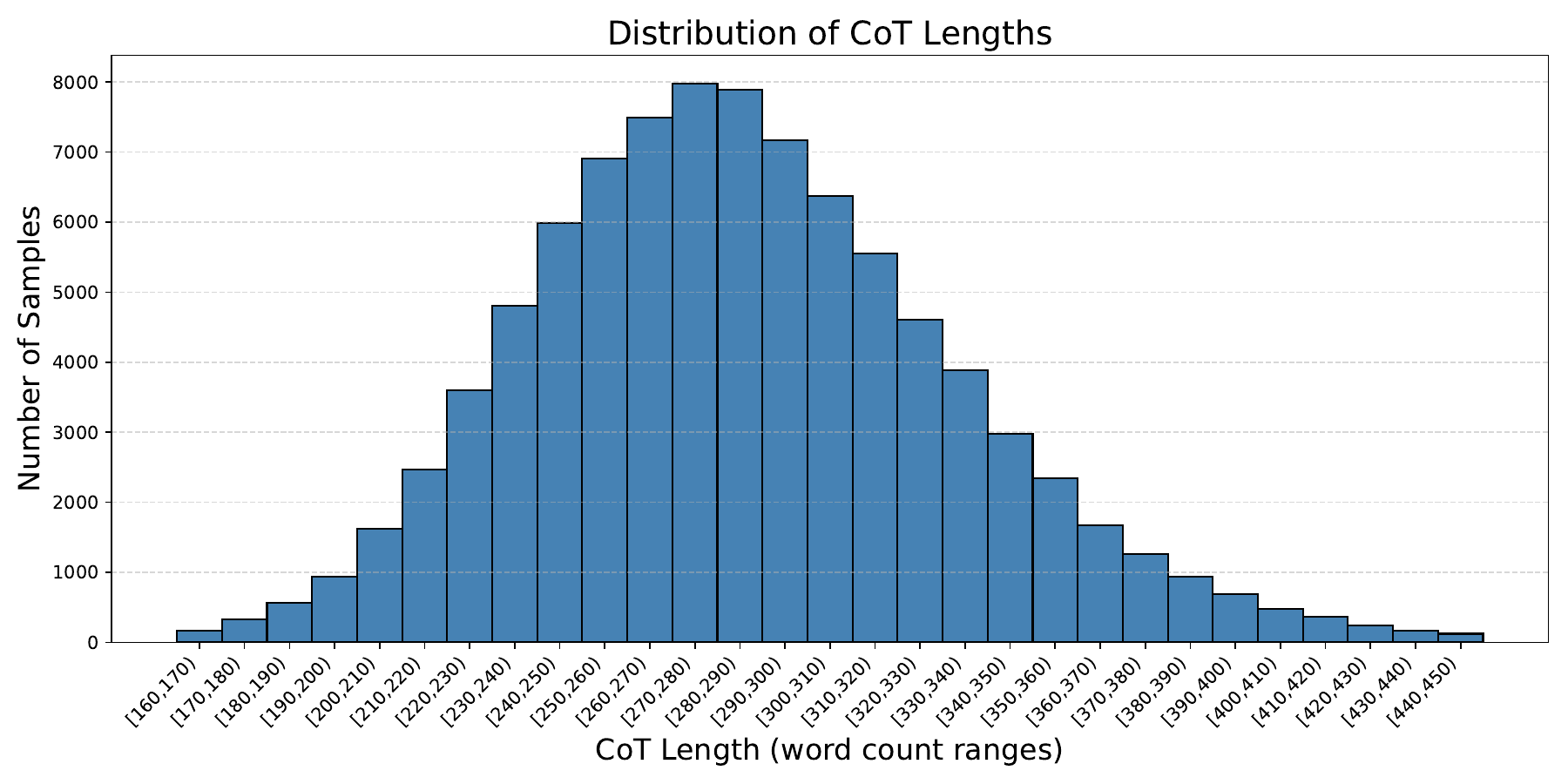}
        \caption{}
        \label{fig:cot_length}
    \end{subfigure}
    \caption{(a) Word cloud of Chain-of-Thought in the R2R-CoT-320k dataset, highlighting frequent visual and spatial reasoning patterns.  (b) Distribution of CoT lengths (in word count), showing a wide and diverse range of reasoning complexity.} % Add an overall caption for the figure
    \label{fig:cot_data}
\end{figure}

To provide a deeper quantitative understanding of our proposed R2R-CoT-320k dataset, we present statistics on CoT content and complexity. As shown in Fig. \ref{fig:word_cloud}, the word cloud reveals frequent reasoning patterns grounded in spatial semantics, such as “doorway,” “current observation,” “hallway,” “turning,” and “goal.” These tokens suggest that the dataset captures rich, step-by-step reasoning tightly aligned with embodied navigation semantics.

Fig. \ref{fig:cot_length}  shows the distribution of CoT lengths, where most reasoning chains fall within the 200–300 word range, but with a long tail reaching beyond 450 words. This indicates that the dataset covers both concise and highly detailed reasoning processes, posing a greater challenge than typical short-form CoT datasets used in static tasks.

Overall, R2R-CoT-320k represents the first large-scale reasoning-augmented dataset for VLN with diverse, high-coverage CoT annotations. It offers a valuable benchmark to study the role of language-based reasoning in long-horizon, partially observable navigation tasks.

\subsection{Navigation Prompts} 
\label{sec:prompt}
We use the following prompt to drive the model to predict navigation actions:

\begin{tcolorbox}[
  colback=gray!15,    % 背景颜色 (e.g., gray!10 for 10% gray, lightgray)
  colframe=black!75,  % 边框颜色 (e.g., black, darkgray)
  boxrule=0.5pt,      % 边框线条粗细
  arc=1mm,            % 边框圆角半径 (0mm for sharp corners)
  % title=Prompt Example, % 如果需要标题，取消注释这一行
]
Imagine you are a robot programmed for navigation tasks. You have been given a video of historical observations: <image>,...,<image> and and current observation: <image>. Your assigned task is: [Instruction]. Analyze this series of images to decide your next move, which could involve turning left or right by a specific degree, moving forward a certain distance, or stop if the task is completed.
\end{tcolorbox}

Among them, [Instruction] is the language instruction given for the current task. For the auxiliary task of CoT-based reasoning, we add “Please provide your step-by-step reasoning process” after the above prompt.

For the Non-CoT Instruction Reasoning, we set the prompt as:
\begin{tcolorbox}[
  colback=gray!15,    % 背景颜色 (e.g., gray!10 for 10% gray, lightgray)
  colframe=black!75,  % 边框颜色 (e.g., black, darkgray)
  boxrule=0.5pt,      % 边框线条粗细
  arc=1mm,            % 边框圆角半径 (0mm for sharp corners)
  % title=Prompt Example, % 如果需要标题，取消注释这一行
]
Assume you are a robot designed for navigation. You are provided with captured image sequences: <image>,...,<image>. Based on this image sequence, please describe the navigation trajectory of the robot.
\end{tcolorbox}

\subsection{More Details About Pre-Think and Post-Think}
\label{sec:more}

To further investigate Test-time Reasoning Collapse (TRC) phenomenon, we introduce special tokens to delineate the reasoning and action prediction components. During training, we assign different loss weights to the reasoning component, as shown in Table~\ref{tab:loss_weight}. We observe that moderately reducing the CoT (Chain-of-Thought) loss weight improves the performance of both Pre-Think and Post-Think. 
However, their performance still lags behind that of No-Think and Aux-Think.

Our results reveal a consistent trend: reducing the CoT loss weight moderately improves performance for both Pre-Think and Post-Think. For example, in the Post-Think setting, decreasing the CoT weight from 1 to 0.1 leads to a +1.8\% SR improvement (from 29.0 to 30.8). This suggests that over-reliance on CoT can introduce noise, potentially due to its misalignment with the suboptimal, off-distribution states encountered during testing.

However, even with carefully tuned weights, both strategies still fall short of No-Think. For instance, No-Think achieves 35.1\% SR, significantly outperforming the best Post-Think variant (30.8\%) and Pre-Think (15.9\%). This persistent gap underscores a deeper issue: while CoT can serve as useful supervision during training, explicitly generating and relying on CoT during testing is inherently brittle in VLN due to compounding errors and distribution shift. This reinforces our finding that VLN agents suffer from reasoning collapse when required to generate structured thoughts in real time within dynamic, partially observable environments.

In summary, despite our extensive efforts to optimize Pre-Think and Post-Think through CoT loss reweighting and architectural adjustments, these strategies fail to match the robustness and effectiveness of CoT-free testing (No-Think). These findings motivate our proposal of Aux-Think, which leverages the strengths of CoT through auxiliary supervision while circumventing the vulnerabilities of test-time reasoning.

\begin{table}[htbp]
\centering
\caption{Experiments on the impact of CoT Loss weight on Pre-Think and Post-Think strategies.}
\label{tab:loss_weight}
\begin{tabular}{c|c|cccc}
\toprule
                              & Weight of CoT Loss    & NE $\downarrow$                  & OSR$\uparrow$                  & SR$\uparrow$                   & SPL$\uparrow$                  \\ \midrule
\multirow{3}{*}{Pre-Think}    & 1                     & 9.23                 & 19.3                 & 11.4                 & 8.6                  \\
                              & 0.1                   & 9.42                 & 28.3                 & 15.9                 & 12.8                 \\
                              & 0.01                  & 8.84                 & 20.6                 & 12.2                 & 9.3                  \\ \midrule
\multirow{3}{*}{Post-Think}   & 1                     & 8.59                 & 35.1                 & 29.0                 & 23.8                     \\
                              & 0.1                   & 8.50                 & 37.7                 & 30.8                 & 24.7                  \\
                              & 0.01                  & 8.25                 & 36.6                 & 29.3                 & 24.9                 \\ \midrule
\multicolumn{1}{l|}{No-Think} & \multicolumn{1}{c|}{-} & \multicolumn{1}{c}{\textbf{7.78}} & \multicolumn{1}{c}{\textbf{43.7}} & \multicolumn{1}{c}{\textbf{35.1}} & \multicolumn{1}{c}{\textbf{30.2}} \\ \bottomrule
\end{tabular}
\end{table}

\end{document}